\documentclass{article}
\usepackage{iclr2027_conference,times}

\usepackage{amsmath,amsfonts,bm}

\newcommand{\newterm}[1]{{\bf #1}}

\def\eqref#1{equation~\ref{#1}}

\def\1{\bm{1}}

\def\ve{{\bm{e}}}

\def\vz{{\bm{z}}}

\DeclareMathAlphabet{\mathsfit}{\encodingdefault}{\sfdefault}{m}{sl}
\SetMathAlphabet{\mathsfit}{bold}{\encodingdefault}{\sfdefault}{bx}{n}

\newcommand{\E}{\mathbb{E}}

\newcommand{\softmax}{\mathrm{softmax}}

\usepackage{tikz}


\usepackage{hyperref}
\hypersetup{
  colorlinks=true,
  linkcolor=orange!50!brown,
  citecolor=teal,
  urlcolor=magenta,
  filecolor=magenta,
}
\usepackage{url}
\usepackage{float}
\usepackage{algorithm}
\usepackage{algpseudocode}
\usepackage{amsmath}
\usepackage{amsthm}
\usepackage{amssymb}
\usepackage{mathtools}
\usepackage{comment}
\usepackage{booktabs}
\usepackage{colortbl}
\usepackage{wrapfig}
\usepackage{graphicx}
\usepackage{tcolorbox}
\usepackage{enumitem} 
\usepackage{subcaption}

\newcommand{\eqp}[1]{Eq.~(\ref{#1})}
\newcommand{\sflm}{$\mathbb{S}$-FLM}
\newcommand{\sflmcont}{Cont.\@}
\newcommand{\sflmtft}{TFT}
\newcommand{\vocab}{\mathcal{V}}
\newcommand{\sph}{\mathbb{S}^{d-1}}
\newcommand{\Logm}[1]{\mathrm{Log}_{#1}}

\newcommand{\ind}{\mathbf{1}}

\theoremstyle{remark}

\usepackage{mdframed}

\surroundwithmdframed[
  backgroundcolor=gray!5,
  linecolor=gray!50,
  linewidth=0.5pt,
  innertopmargin=6pt,
  innerbottommargin=6pt
]{proposition}

\surroundwithmdframed[
  backgroundcolor=gray!5,
  linecolor=gray!50,
  linewidth=0.5pt,
  innertopmargin=6pt,
  innerbottommargin=6pt
]{assumption}

\surroundwithmdframed[
  backgroundcolor=gray!5,
  linecolor=gray!50,
  linewidth=0.5pt,
  innertopmargin=6pt,
  innerbottommargin=6pt
]{corollary}

\title{BCT: Bayes-Consistent Tempering for\\ Flow Language Models}

\author{Daisuke Oba\textsuperscript{1}\thanks{Corresponding author. Email: \texttt{daisuke.oba@nlp.comp.isct.ac.jp}} \quad Hiroki Furuta\quad Naoaki Okazaki\textsuperscript{1,2,3} \\
\textsuperscript{1}Institute of Science Tokyo\quad \textsuperscript{2}AIST \quad  \textsuperscript{3}NII LLMC \\
}

\iclrfinalcopy

\begin{document}

\maketitle

\lhead{Preprint}

\begin{abstract} 
  Temperature scaling is typically applied as a post-hoc transformation of model probabilities. In posterior-induced flow language models, this creates two mismatches. First, posterior tempering implicitly changes both the endpoint distribution and the distribution of intermediate noisy states, while the denoiser remains trained on the original states. Second, the tempered posterior weights are combined with transport dynamics derived for the untempered flow. These mismatches are reflected empirically: lowering temperature improves GSM8K performance much less reliably in flow LMs than in discrete diffusion LMs. We introduce \textbf{Bayes-Consistent Tempering (BCT)}, which makes temperature explicit in both the distributions and dynamics of the generative flow. Its learned realization trains the denoiser on tempered intermediate states and tempers the endpoint distribution using the corresponding tempered dynamics. We also derive training-free variants for conventionally trained checkpoints. Across hyperspherical flow LMs with two denoiser architectures and a von Mises--Fisher flow LM, BCT improves GSM8K generation across broad temperature ranges and achieves the strongest tempering performance. The gains also extend to Sudoku, demonstrating benefits on tightly constrained generation. At very low temperatures, the learned realization is also markedly more robust than the training-free variants, supporting temperature as a property of the generative flow across training and inference rather than a post-hoc probability adjustment. Code and model weights are available at:
  \par{\centering\url{https://daioba.github.io/bct}.\par}
\end{abstract}

\begin{figure*}[h]
    \centering
    \includegraphics[width=0.9\textwidth]{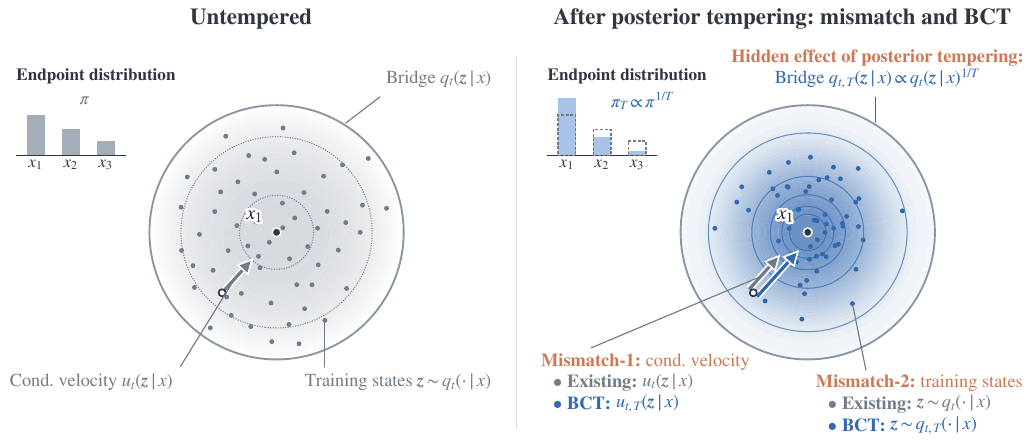}
    \caption{
        \textbf{Conceptual overview of Bayes-Consistent Tempering (BCT).}
        Posterior tempering implicitly changes both the endpoint distribution
        and the conditional bridge, while standard sampling retains the
        original conditional velocities and denoiser training states.
        BCT restores consistency between the bridge and sampling dynamics,
        and its learned realization additionally aligns training with
        the tempered bridge.
    }
    \label{fig:bct_concept}
\end{figure*}
\clearpage
\section{Introduction}
\label{sec:intro}
Temperature scaling is one of the simplest controls for language generation \citep{ackley1985learning,holtzman2020curious}. In autoregressive models, dividing next-token logits by \(T<1\) directly sharpens the distribution from which the next token is sampled. In many flow language models for discrete data, however, generation is driven by a denoising posterior: a model predicts the clean endpoint token from an intermediate noisy state, and this posterior induces the velocity that transports the state toward token embeddings \citep{stark2024dirichlet,fmlm2026,hypersphericalflows2026}. Applying temperature only to this posterior creates two mismatches. First, by Bayes' rule, posterior tempering implicitly changes not only the endpoint distribution but also the \emph{conditional bridge}---the time-indexed distribution of intermediate noisy states given each endpoint. The denoiser, however, is trained on states from the untempered bridge, so tempering changes the states encountered at inference without a corresponding change in training. Second, tempering changes the posterior weights over endpoints but leaves their conditional velocities unchanged, even though those velocities are designed for the original, untempered bridges. Thus, temperature changes the implied generative distribution without consistently changing how the model is trained or sampled.

This issue is also visible empirically. On GSM8K~\citep{cobbe2021training}, prior work reports that flow language models gain little from lowering the temperature, in contrast to discrete diffusion models such as MDLM~\citep{sahoo2024simple} and Duo~\citep{sahoo2025the}, whose accuracy improves substantially under low-temperature decoding~\citep{hypersphericalflows2026}. This discrepancy motivates us to ask whether temperature is being incorporated correctly into posterior-induced flows.

We introduce \textbf{Bayes-Consistent Tempering (BCT)}, which treats temperature as part of the generative flow rather than as a post-hoc transformation of the denoising posterior (Sec.~\ref{sec:bct}). 
Its learned realization trains the denoiser on states from a tempered conditional bridge and, at inference, applies the Bayes correction needed to temper the endpoint distribution while preserving that tempered bridge, together with the dynamics induced by the same bridge. We call the training component \textbf{Tempered-Flow Training (TFT)} and the corresponding sampler \textbf{BCT-L}. 

We further investigate how much inference-time consistency can be recovered without retraining through two variants for existing checkpoints: \textbf{BCT-P} tempers both the endpoint distribution and bridge with matched dynamics, whereas \textbf{BCT-E} tempers only the endpoint distribution, retaining the original bridge and dynamics (Sec.~\ref{sec:posthoc-bct}).

For posterior-induced flow language models, BCT is not merely a conceptual framework: applying it requires deriving the tempered bridge and its corresponding sampling dynamics, and these quantities are tractable in concrete models. For \sflm{}~\citep{hypersphericalflows2026}, for example, the hyperspherical geometry lets us derive both explicitly. The same principle applies to SFCD~\citep{sphericalflows2026}, which uses a von Mises--Fisher bridge with its own analytic tempering rule and corresponding dynamics. Thus, BCT is not specific to a particular denoiser architecture; its construction is determined by the bridge that defines the flow.

Across hyperspherical flow LMs with two denoiser architectures (S-arch/DiT)~\citep{hypersphericalflows2026} and a von Mises--Fisher flow LM~\citep{sphericalflows2026}, BCT improves GSM8K over broad temperature ranges (Sec.~\ref{sec:experiments}). 
BCT-L is particularly robust at very low temperatures, outperforming the training-free variants and highlighting the benefit of temperature-aware training. Gains also extend to Sudoku~\citep{sudokuben,sudokukim}, where standard tempering does not benefit from lower temperatures, suggesting applicability to constrained generation.
\section{Preliminaries}
\label{sec:prelim}
\subsection{Posterior-induced flows}
We consider a single token position with finite vocabulary $\vocab$ and endpoint distribution $\pi(x)$. Let $\vz$ denote the continuous latent state and $t\in[0,1]$ run from noise to data. For each endpoint token $x$, a \newterm{conditional bridge} $q_t(\vz\mid x)$ describes the distribution of intermediate states at time $t$, starting from a token-independent source $p_0$ and converging toward its embedding $\ve_x$. All bridge densities are defined with respect to a fixed reference measure $\mu$ on the latent space. A corresponding \newterm{conditional velocity} $u_t(\vz\mid x)$ transports states so that their distribution follows $q_t(\cdot\mid x)$ over time; formally, $q_t$ and $u_t$ satisfy the continuity equation.

Bayes' rule gives the denoising posterior
\begin{equation}
r_t(x\mid\vz)
=
\frac{\pi(x)\,q_t(\vz\mid x)}
{\sum_{y\in\vocab}\pi(y)\,q_t(\vz\mid y)},
\label{eq:bayes}
\end{equation}
and the generative velocity is its posterior-weighted average of the conditional velocities,
\begin{equation}
u_t(\vz)
=
\sum_{x\in\vocab}
r_t(x\mid\vz)\,u_t(\vz\mid x).
\label{eq:marg-vel}
\end{equation}
A denoiser $r^\eta_t(x\mid\vz)$ is trained to approximate $r_t$ and replaces it in \eqp{eq:marg-vel} at generation time. We call flows constructed in this way \newterm{posterior-induced flows}.

\subsection{Standard tempering}

Writing $\beta=1/T$, standard temperature scaling replaces the denoising posterior by
\begin{equation}
r_t^T(x\mid\vz)
\propto
r_t(x\mid\vz)^\beta,
\qquad
u_t^T(\vz)
=
\sum_x r_t^T(x\mid\vz)\,u_t(\vz\mid x),
\label{eq:std-temp}
\end{equation}
thus changing the posterior weights while leaving the conditional velocities unchanged. In practice, this is implemented by dividing the denoiser logits by $T$.

Using \eqp{eq:bayes},
\begin{equation}
r_t(x\mid\vz)^\beta
\propto
\pi(x)^\beta q_t(\vz\mid x)^\beta,
\label{eq:temp-decomp}
\end{equation}
so posterior tempering acts algebraically on both the endpoint factor and the bridge likelihood. Define
\begin{equation}
\pi_T(x)\propto\pi(x)^\beta,
\qquad
q_{t,T}(\vz\mid x)
=
\frac{q_t(\vz\mid x)^\beta}{Z_{t,T}(x)},
\qquad
Z_{t,T}(x)
=
\int q_t(\vz\mid x)^\beta\,d\mu(\vz).
\label{eq:tempered-pair}
\end{equation}
The posterior induced by $(\pi_T,q_{t,T})$ differs from standard posterior tempering only by the token-dependent normalizer $Z_{t,T}(x)$. Hence the two coincide whenever $Z_{t,T}(x)$ is independent of $x$, as for the symmetric bridge families studied in this work.

For clarity, we present all derivations for a single token position. For sequences, the same decomposition applies position-wise, with $\pi(x)$ replaced by the context-conditioned endpoint distribution induced by the remaining positions; see Appendix~\ref{app:sequences}.

\section{Bayes-Consistent Tempering}
\label{sec:bct}

Once an endpoint distribution and a conditional bridge are specified, the posterior weights and sampling dynamics should correspond to that same generative model. In particular, for the tempered pair $(\pi_T,q_{t,T})$, the sampler should use its denoising posterior, conditional velocities $u_{t,T}(\vz\mid x)$ that transport $q_{t,T}$, and the source $q_{0,T}$ of that bridge.
However, standard tempering does not generally satisfy this alignment: while its posterior corresponds to the tempered pair under the condition described in Section~\ref{sec:prelim}, it continues to use the conditional velocities $u_t(\vz\mid x)$ of the original bridge. 

In contrast, our proposal, \newterm{Bayes-Consistent Tempering (BCT)}, treats the endpoint distribution, conditional bridge, and sampling dynamics as components of the same generative flow. 
We first develop its learned realization, which also aligns the distribution of states seen during training (Sec.~\ref{sec:learned-bct}), and then derive post-hoc variants for conventionally trained models (Sec.~\ref{sec:posthoc-bct}).

\subsection{Learned BCT}
\label{sec:learned-bct}

The learned realization of BCT, \newterm{BCT-L}, aligns training with the tempered bridge used at inference. 
We first train the denoiser on intermediate states drawn from $q_{t,T}$, rather than from the original bridge $q_t$. 
Given a training distribution $p(T)$ over temperatures, \newterm{Tempered-Flow Training (TFT)} minimizes the usual denoising loss,
\begin{equation}
\mathcal{L}(\eta)
=
\E_{T\sim p(T),\,t,\,x\sim\pi,\,\vz\sim q_{t,T}(\cdot\mid x)}
\left[-\log r^\eta_{t,T}(x\mid\vz)\right],
\label{eq:tft}
\end{equation}
where the denoiser is conditioned on $T$. 
Since the endpoint token is still drawn from $\pi$ while its intermediate state is drawn from tempered bridge $q_{t,T}$, the Bayes-optimal denoiser is
\begin{equation}
r_{t,T}(x\mid\vz)
=
\frac{\pi(x)\,q_{t,T}(\vz\mid x)}
{\sum_y \pi(y)\,q_{t,T}(\vz\mid y)}.
\label{eq:tft-posterior}
\end{equation}
Thus, TFT incorporates the tempered bridge into the learned posterior, but does not itself temper the endpoint distribution.

At inference, BCT targets the pair $(\pi_T,q_{t,T})$. 
Its tempered posterior can be written directly in terms of the TFT posterior as
\begin{equation}
r^{\mathrm{BCT\text{-}L}}_{t,T}(x\mid\vz)
\propto
r_{t,T}(x\mid\vz)^\beta
q_{t,T}(\vz\mid x)^{1-\beta}.
\label{eq:bctl-posterior}
\end{equation}
The second factor is the Bayes correction that tempers the endpoint while preserving the tempered bridge already represented by the denoiser. In practice, learned posteriors are approximate, so we allow the strength of this correction to be selected on held-out data:
\begin{equation}
\ell^{\mathrm{BCT\text{-}L}}_x
=
\beta\,\ell^\eta_x
+
\lambda(1-\beta)\log q_{t,T}(\vz\mid x),
\qquad
0\le\lambda\le1,
\label{eq:bctl-logits}
\end{equation}
where one $\lambda$ is selected per trained model and shared across temperatures. 

Finally, BCT-L uses the tempered conditional velocity $u_{t,T}(\vz\mid x)$ that transports the corresponding tempered bridge $q_{t,T}$:
\begin{equation}
u^{\mathrm{BCT\text{-}L}}_{t,T}(\vz)
=
\sum_x
r^{\mathrm{BCT\text{-}L}}_{t,T}(x\mid\vz)\,
u_{t,T}(\vz\mid x),
\label{eq:bctl-velocity}
\end{equation}
with sampling initialized from the tempered source $q_{0,T}$. 
The form of $u_{t,T}$ is model-dependent; we derive it for the flow constructions studied here in Sec.~\ref{sec:bct-instantiations}.
Thus, TFT and BCT-L align the intermediate states seen during training, the endpoint distribution, and the sampling dynamics within the same generative flow.

\subsection{Post-hoc BCT Variants for Conventionally Trained Models}
\label{sec:posthoc-bct}

We next ask how much of BCT can be recovered after a model has already been trained on the original bridge $q_t$. We consider two complementary post-hoc variants---\newterm{BCT-P} and \newterm{BCT-E}.

\paragraph{Path tempering (BCT-P).}
BCT-P retains posterior tempering, which implicitly tempers both the endpoint distribution and the bridge, but replaces the original conditional dynamics with those of the tempered bridge. Its posterior is
\begin{equation}
r^{\mathrm{BCT-P}}_{t,T}(x\mid\vz)
\propto
\frac{r_t(x\mid\vz)^\beta}{Z_{t,T}(x)},
\label{eq:bctp-posterior}
\end{equation}
and the corresponding velocity is
\begin{equation}
u^{\mathrm{BCT-P}}_{t,T}(\vz)
=
\sum_x
r^{\mathrm{BCT-P}}_{t,T}(x\mid\vz)\,
u_{t,T}(\vz\mid x).
\label{eq:bctp-velocity}
\end{equation}
When $Z_{t,T}(x)$ is independent of $x$, BCT-P uses exactly the same posterior weights as standard tempering; the difference is that its conditional velocities are matched to the tempered bridge. Unlike BCT-L, however, its denoiser was trained only on states from the original bridge $q_t$, leaving a train--inference mismatch in the intermediate states.

\paragraph{Endpoint tempering (BCT-E).}
BCT-E avoids tempering the bridge altogether. 
It keeps untempered bridge and conditional dynamics, and tempers only the endpoint distribution $\pi$. 
By Bayes' rule, its posterior can be written using the original denoiser as
\begin{equation}
r^{\mathrm{BCT-E}}_{t,T}(x\mid\vz)
\propto
r_t(x\mid\vz)^\beta
q_t(\vz\mid x)^{1-\beta},
\label{eq:bcte-posterior}
\end{equation}
or, in logits,
\begin{equation}
\ell^{\mathrm{E}}_x
=
\beta\,\ell_x
+
(1-\beta)\log q_t(\vz\mid x).
\label{eq:bcte-logits}
\end{equation}
These weights are combined with the original conditional velocities $u_t(\vz\mid x)$.

Both variants require no retraining. 
BCT-P restores consistency by matching the sampling dynamics to the tempered bridge, 
whereas BCT-E restores consistency by keeping the bridge and sampling dynamics untempered.

\subsection{Instantiations}
\label{sec:bct-instantiations}

We instantiate BCT on two posterior-induced flow constructions with different conditional bridges.

\textbf{Hyperspherical flow.}
S-FLM~\citep{hypersphericalflows2026} is based on hyperspherical flow, which places token embeddings and latents on the unit sphere and transports a uniform source toward each endpoint along geodesics. Let
$\theta_x(\vz)=\arccos\langle \vz,\ve_x\rangle$ and $a_t=1-\alpha_t$ denote the remaining fraction of the geodesic at time $t$. Its conditional bridge has density
\begin{equation}
q_t(\vz\mid x)
\propto
\frac{\sin^{d-2}(\theta_x/a_t)}
{\sin^{d-2}\theta_x}
\;\ind\{\theta_x<a_t\pi\},
\label{eq:sflm-bridge}
\end{equation}
with respect to the spherical surface measure. This density gives the bridge likelihood needed by BCT-E and TFT. Rotational symmetry also makes $Z_{t,T}(x)$ independent of $x$, so BCT-P uses the same posterior weights as standard tempering.

The tempered bridge $q_{t,T}$ remains rotationally symmetric around $\ve_x$. Its corresponding conditional velocity therefore follows the same geodesic direction as the original S-FLM velocity, with a temperature-dependent change in speed,
\begin{equation}
u_{t,T}(\vz\mid x)
=
w_{t,T}(\theta_x)\,u_t(\vz\mid x),
\label{eq:sflm-tempered-velocity}
\end{equation}
where $w_{t,T}$ is obtained from a one-dimensional continuity equation; we give its derivation in Appendix~\ref{app:velocity}. The same rotational symmetry reduces exact sampling from $q_{t,T}$ to a one-dimensional transformation of the polar angle, allowing TFT to draw training states directly from the tempered bridge (Appendix~\ref{app:corruption}).

\textbf{von Mises--Fisher flow.}
\cite{sphericalflows2026} use a von Mises--Fisher conditional bridge,
\begin{equation}
q_t(\vz\mid x)
\propto
\exp\!\left(\kappa_t\langle \ve_x,\vz\rangle\right).
\label{eq:vmf-bridge}
\end{equation}
Tempering stays within the same family:
\begin{equation}
q_{t,T}(\vz\mid x)
\propto
\exp\!\left(\frac{\kappa_t}{T}\langle \ve_x,\vz\rangle\right),
\label{eq:vmf-tempered}
\end{equation}
so BCT simply replaces $\kappa_t$ by $\kappa_t/T$ and uses the corresponding conditional dynamics (Appendix~\ref{app:sfcd}). TFT likewise draws training states directly from the vMF distribution with concentration $\kappa_t/T$. The normalizer is independent of $x$, and BCT-P therefore again shares its posterior weights with standard tempering.

For von Mises--Fisher flow, the tempered bridge evidence
$g_{t,T}(x,\vz)=(\kappa_t/T)\langle\ve_x,\vz\rangle$
is available analytically. For TFT, we therefore include this term explicitly in the denoiser logits,
$\ell^\eta_x=a_{\eta,x}+g_{t,T}(x,\vz)$,
and learn only the remaining component $a_{\eta,x}$. This is a parameterization of the same TFT posterior and BCT-L sampler, not a separate training method.

\section{Experiments}
\label{sec:experiments}

\subsection{Experimental Setup}
\label{sec:exp-setup}

\textbf{Main setting.}
Our main experiments evaluate BCT on GSM8K~\citep{cobbe2021training} with \sflm{}~\citep{hypersphericalflows2026}.
Following its setup, we adopt the TinyGSM protocol~\citep{liu2023tinygsm}: for each of the 1319 test problems, the model generates a Python program, and the problem counts as solved if executing the program returns the reference answer.
We use its released TinyGSM checkpoints for two denoiser backbones: S-arch, a Transformer whose intermediate activations lie on the hypersphere, and the standard Diffusion Transformer~\citep[DiT;][]{dit}.
We refer to either checkpoint as \sflm{} and specify the backbone where relevant.
The two backbones share the bridge, the posterior-induced velocity, and all samplers, so comparing them varies only the denoiser architecture.

\textbf{Additional settings.}
Two additional experiments test whether our conclusions generalize to other settings (Sections~\ref{sec:exp-sudoku} and~\ref{sec:exp-sfcd}).
First, we change the task while keeping the hyperspherical flow, using Sudoku-Hard~\citep{sudokukim,sudokuben}, which requires tightly constrained generation.
Since no Sudoku checkpoint is released, we train a DiT-based \sflm{} with its released training recipe (Appendix~\ref{app:sudoku}).
Second, we change the flow structure while keeping the task: we apply BCT to SFCD~\citep{sphericalflows2026}, which uses a von Mises--Fisher bridge instead of the geodesic bridge of \sflm{}, and evaluate its released TinyGSM checkpoint on the same GSM8K problems (Appendix~\ref{app:sfcd}).
Unless noted otherwise in their sections, both follow the protocol below.

\textbf{Samplers.}
All samplers use the exact posterior-induced velocity with 1024 function evaluations (NFE) and otherwise follow the released sampler; they differ only in the posterior weights and conditional velocities.
We compare standard tempering \eqp{eq:std-temp}, BCT-E \eqp{eq:bcte-logits}, BCT-P \eqp{eq:bctp-velocity}, and BCT-L, which is described below.
As a reference, we report the Top-1 velocity~\citep{hypersphericalflows2026}, which moves only toward the most likely token; it is invariant to temperature and is the $T\to0$ limit of standard tempering.
We evaluate $T\in\{1,0.8,0.6,0.5,0.4,0.3,0.2,0.1\}$ in all experiments.

\textbf{TFT and BCT-L.}
For each backbone, we continue training the released checkpoint with TFT \eqp{eq:tft} for 2{,}000 updates, 0.8\% of its 250k pretraining updates, and refer to the resulting model as \sflmtft{}.
The temperature distribution $p(T)$ draws $T=1$ with probability 0.2, $T=0.1$ with probability 0.1, and otherwise $\beta\sim U(1.25,5)$, so it covers every evaluated temperature.
As a matched training control, \sflmcont{} continues training for the same 2{,}000 updates with the original recipe.
All models are evaluated with EMA weights.
$\lambda$ in \eqp{eq:bctl-logits} is selected once per model on TinyGSM validation problems and shared across temperatures (Appendix~\ref{app:practical}).

\begin{figure}[t]
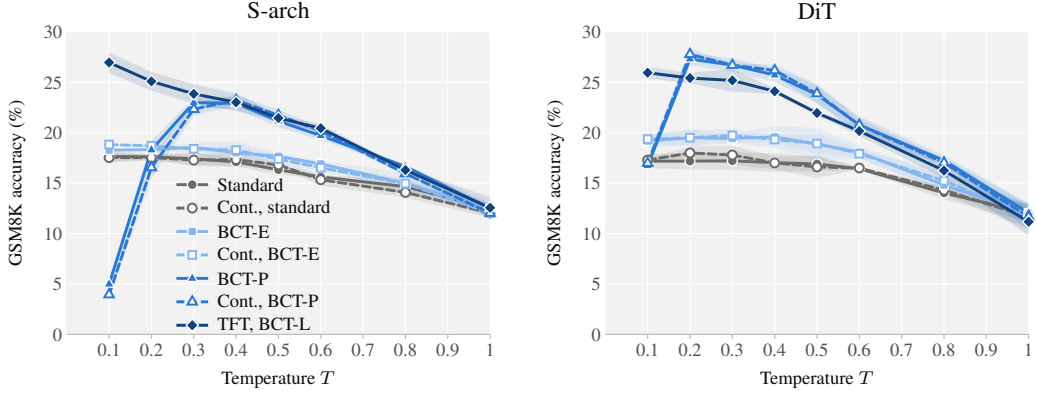

\centering
\begin{minipage}[t]{0.49\linewidth}
\centering
\input{figures/fig_main_sarch}
\end{minipage}\hfill
\begin{minipage}[t]{0.49\linewidth}
\centering
\input{figures/fig_main_sdit}
\end{minipage}
\caption{GSM8K accuracy against the sampling temperature on the two backbones (NFE $=1024$). Lines are the mean over three generation seeds and bands $\pm$ one sample standard deviation. Curves without a prefix use the released checkpoint; Cont.\ (dashed) is the matched continuation. At $T=1$ every BCT correction is the identity. Values in Table~\ref{tab:main}.}
\label{fig:main}
\end{figure}

\begin{table}[t]
\caption{GSM8K accuracy (\%), mean $\pm$ sample standard deviation over three generation seeds (1319 problems, NFE $=1024$).}
\label{tab:main}
\centering
\small
\setlength{\tabcolsep}{4pt}
\begin{tabular}{@{}lccccccc@{}}
\toprule
& \sflmtft{} & \multicolumn{3}{c}{\sflmcont{}} & \multicolumn{3}{c}{\sflm{}} \\
\cmidrule(lr){2-2}\cmidrule(lr){3-5}\cmidrule(l){6-8}
$T$ & BCT-L & Standard & BCT-E & BCT-P & Standard & BCT-E & BCT-P \\
\midrule
\rowcolor{black!8}\multicolumn{8}{c}{\emph{S-arch}} \\
1 & 12.6\,{\scriptsize$\pm$1.1} & 12.0\,{\scriptsize$\pm$0.5} & 12.0\,{\scriptsize$\pm$0.5} & 12.0\,{\scriptsize$\pm$0.5} & 12.5\,{\scriptsize$\pm$0.5} & 12.5\,{\scriptsize$\pm$0.5} & 12.5\,{\scriptsize$\pm$0.5} \\
0.8 & 16.3\,{\scriptsize$\pm$0.2} & 14.1\,{\scriptsize$\pm$0.5} & 15.0\,{\scriptsize$\pm$0.5} & 15.9\,{\scriptsize$\pm$0.2} & 14.7\,{\scriptsize$\pm$0.7} & 15.0\,{\scriptsize$\pm$1.1} & 16.7\,{\scriptsize$\pm$0.2} \\
0.6 & 20.4\,{\scriptsize$\pm$0.4} & 15.3\,{\scriptsize$\pm$0.6} & 16.5\,{\scriptsize$\pm$0.4} & 20.2\,{\scriptsize$\pm$0.3} & 15.6\,{\scriptsize$\pm$0.2} & 16.9\,{\scriptsize$\pm$0.3} & 19.7\,{\scriptsize$\pm$0.5} \\
0.5 & 21.5\,{\scriptsize$\pm$0.3} & 16.8\,{\scriptsize$\pm$0.2} & 17.4\,{\scriptsize$\pm$0.4} & 21.7\,{\scriptsize$\pm$0.6} & 16.3\,{\scriptsize$\pm$0.5} & 17.7\,{\scriptsize$\pm$0.3} & 21.2\,{\scriptsize$\pm$0.6} \\
0.4 & 23.0\,{\scriptsize$\pm$0.9} & 17.3\,{\scriptsize$\pm$0.7} & 18.3\,{\scriptsize$\pm$0.5} & 23.2\,{\scriptsize$\pm$0.5} & 17.1\,{\scriptsize$\pm$0.5} & 18.0\,{\scriptsize$\pm$1.1} & 22.9\,{\scriptsize$\pm$0.6} \\
0.3 & 23.9\,{\scriptsize$\pm$1.0} & 17.3\,{\scriptsize$\pm$0.6} & 18.4\,{\scriptsize$\pm$0.2} & 22.3\,{\scriptsize$\pm$0.6} & 17.4\,{\scriptsize$\pm$0.6} & 18.5\,{\scriptsize$\pm$0.6} & 23.0\,{\scriptsize$\pm$0.5} \\
0.2 & 25.1\,{\scriptsize$\pm$0.9} & 17.5\,{\scriptsize$\pm$0.3} & 18.7\,{\scriptsize$\pm$0.1} & 16.6\,{\scriptsize$\pm$1.1} & 17.6\,{\scriptsize$\pm$0.3} & 18.3\,{\scriptsize$\pm$0.1} & 18.3\,{\scriptsize$\pm$0.9} \\
0.1 & 27.0\,{\scriptsize$\pm$1.0} & 17.5\,{\scriptsize$\pm$0.5} & 18.8\,{\scriptsize$\pm$0.1} & 3.9\,{\scriptsize$\pm$0.5} & 17.6\,{\scriptsize$\pm$0.5} & 18.3\,{\scriptsize$\pm$0.1} & 5.0\,{\scriptsize$\pm$0.4} \\
\midrule
\rowcolor{black!8}\multicolumn{8}{c}{\emph{DiT}} \\
1 & 11.2\,{\scriptsize$\pm$1.3} & 11.7\,{\scriptsize$\pm$1.2} & 11.7\,{\scriptsize$\pm$1.2} & 11.7\,{\scriptsize$\pm$1.2} & 12.0\,{\scriptsize$\pm$1.0} & 12.0\,{\scriptsize$\pm$1.0} & 12.0\,{\scriptsize$\pm$1.0} \\
0.8 & 16.2\,{\scriptsize$\pm$0.8} & 14.3\,{\scriptsize$\pm$0.4} & 15.2\,{\scriptsize$\pm$1.0} & 17.0\,{\scriptsize$\pm$0.4} & 14.0\,{\scriptsize$\pm$0.6} & 14.8\,{\scriptsize$\pm$0.9} & 17.3\,{\scriptsize$\pm$0.5} \\
0.6 & 20.2\,{\scriptsize$\pm$0.5} & 16.5\,{\scriptsize$\pm$0.1} & 17.9\,{\scriptsize$\pm$0.4} & 20.8\,{\scriptsize$\pm$0.6} & 16.4\,{\scriptsize$\pm$0.1} & 18.1\,{\scriptsize$\pm$0.6} & 20.8\,{\scriptsize$\pm$0.9} \\
0.5 & 22.0\,{\scriptsize$\pm$0.2} & 16.6\,{\scriptsize$\pm$1.1} & 18.9\,{\scriptsize$\pm$1.0} & 23.9\,{\scriptsize$\pm$0.9} & 16.9\,{\scriptsize$\pm$0.7} & 18.9\,{\scriptsize$\pm$1.6} & 23.8\,{\scriptsize$\pm$0.9} \\
0.4 & 24.1\,{\scriptsize$\pm$0.3} & 17.0\,{\scriptsize$\pm$0.6} & 19.3\,{\scriptsize$\pm$0.3} & 26.2\,{\scriptsize$\pm$0.6} & 17.0\,{\scriptsize$\pm$0.9} & 19.6\,{\scriptsize$\pm$1.1} & 25.7\,{\scriptsize$\pm$0.7} \\
0.3 & 25.2\,{\scriptsize$\pm$1.1} & 17.8\,{\scriptsize$\pm$0.9} & 19.7\,{\scriptsize$\pm$0.6} & 26.7\,{\scriptsize$\pm$0.2} & 17.2\,{\scriptsize$\pm$0.9} & 19.4\,{\scriptsize$\pm$0.5} & 26.7\,{\scriptsize$\pm$0.7} \\
0.2 & 25.4\,{\scriptsize$\pm$0.5} & 18.0\,{\scriptsize$\pm$0.7} & 19.5\,{\scriptsize$\pm$0.7} & 27.8\,{\scriptsize$\pm$0.6} & 17.2\,{\scriptsize$\pm$0.8} & 19.5\,{\scriptsize$\pm$0.8} & 27.3\,{\scriptsize$\pm$0.6} \\
0.1 & 26.0\,{\scriptsize$\pm$0.6} & 17.3\,{\scriptsize$\pm$0.2} & 19.4\,{\scriptsize$\pm$0.6} & 17.0\,{\scriptsize$\pm$0.6} & 17.2\,{\scriptsize$\pm$0.7} & 19.2\,{\scriptsize$\pm$0.6} & 16.7\,{\scriptsize$\pm$0.4} \\
 
\bottomrule
\end{tabular}
\end{table}

\subsection{Main Results}
\label{sec:exp-main}

Figure~\ref{fig:main} and Table~\ref{tab:main} show the results for the hyperspherical flow, whose bridge moves latents along geodesics on the sphere (Section~\ref{sec:bct-instantiations}), with both denoiser backbones.
Since TFT continues a pretrained model rather than training from scratch, the fair baseline for BCT-L is Continual Training (\sflmcont{}), which receives the same additional update steps with the original recipe; the two differ only in whether temperature enters training.
The continuation itself changes little: standard tempering on \sflmcont{} stays within one point of the released checkpoint at every temperature, and the training-free variants behave alike on the two checkpoints.

\textbf{BCT-L makes low-temperature decoding effective.}
With conventional training, lowering the temperature helps little: standard tempering stays at 17--18\% for every $T\le0.4$ on both backbones, close to the Top-1 reference, its $T\to0$ limit.
BCT-L instead improves steadily as the temperature decreases, reaching $27.0\pm1.0\%$ on S-arch and $26.0\pm0.6\%$ on DiT at $T=0.1$, and it is more accurate than standard tempering at every $T<1$.

\textbf{Matching the dynamics at inference gives large gains at moderate temperatures.}
Without training, BCT-P already improves substantially on standard tempering, raising its best accuracy by about 6 points on S-arch and 10 points on DiT, and at moderate temperatures it is comparable to BCT-L, slightly ahead on DiT.

\textbf{Strong tempering requires training on the tempered bridge.}
Under strong tempering, however, BCT-P collapses and falls below standard tempering at $T=0.1$ on both backbones.
This is consistent with its train--inference mismatch (Section~\ref{sec:posthoc-bct}): the latents follow the tempered bridge, on whose states the denoiser was not trained, and the mismatch grows as $T$ decreases.
BCT-L, whose denoiser is trained on the tempered bridge, has no such mismatch and is the most accurate sampler at $T=0.1$.
The results at $T=0.1$ thus particularly highlight the importance of removing the mismatch between the intermediate states seen in training and those visited during sampling.

\textbf{Keeping the bridge is stable but gains less.}
BCT-E keeps the bridge on which the denoiser was trained and therefore does not degrade at low temperature, but it improves only modestly on standard tempering and stays below BCT-L at every $T<1$, increasingly so as the temperature decreases.


\subsection{Gains Beyond Reduced Diversity}
\label{sec:exp-passk}

\begin{figure}[t]
\centering
\input{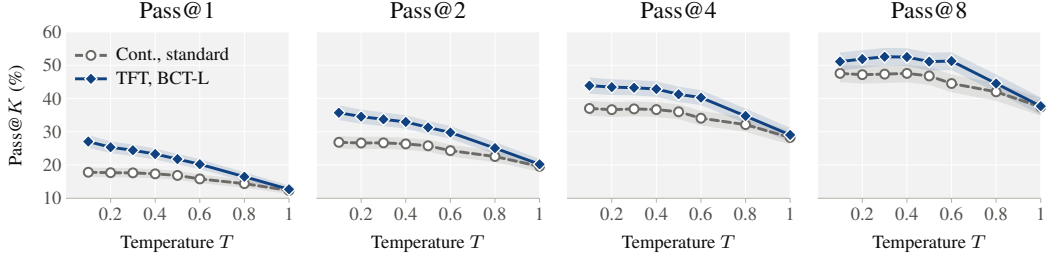}
\caption{Pass@$K$ on GSM8K (S-arch, NFE $=1024$), estimated from eight generations per problem. Bands are 95\% bootstrap intervals over problems; the panels share the vertical axis.}
\label{fig:passk}
\end{figure}

Lowering the temperature concentrates the sampler, so the low-temperature gain of BCT-L could merely trade sample diversity for single-sample accuracy \citep{chen2021codex}, which a less concentrated sampler would recover with more samples.
We test this on S-arch by comparing BCT-L with standard tempering on \sflmcont{} in Pass@$K$ for $K\in\{1,2,4,8\}$, estimated from eight generations per problem with the unbiased estimator of \citet{chen2021codex} (Figure~\ref{fig:passk}; details in Appendix~\ref{app:passk}).

For every $T<1$ and $K\le8$, BCT-L has the higher Pass@$K$, and the 95\% paired bootstrap interval of the difference lies above zero.
More samples raise Pass@$K$ for both samplers, but not enough for standard tempering to catch up.
The low-temperature gain of BCT-L is therefore not explained by reduced diversity alone.

\subsection{Beyond GSM8K: Sudoku-Hard}
\label{sec:exp-sudoku}

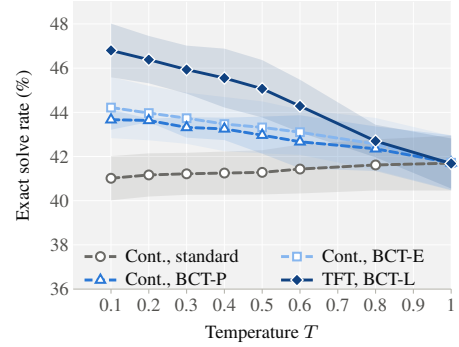
\begin{wrapfigure}[16]{r}{0.47\linewidth}
\vspace{-12pt}
\centering
\begin{tikzpicture}[x=1cm,y=1cm,font=\scriptsize]
\definecolor{vzBlue}{HTML}{2A78D6}
\definecolor{vzOrange}{HTML}{EB6834}
\definecolor{vzAqua}{HTML}{1BAF7A}
\definecolor{vzBase}{HTML}{6B6A66}
\definecolor{vzMuted}{HTML}{9A9892}
\definecolor{vzGrid}{HTML}{E1E0D9}
\definecolor{vzAxis}{HTML}{B5B4AC}
\definecolor{vzInk}{HTML}{0B0B0B}
\definecolor{vzInkTwo}{HTML}{52514E}
\definecolor{vzViolet}{HTML}{4A3AA7}
\definecolor{vzT4}{HTML}{86B6EF}
\definecolor{vzT2}{HTML}{2A78D6}
\definecolor{vzT1}{HTML}{104281}
\definecolor{vzBg}{HTML}{F2F2F2}
\fill[vzBg] (0,0) rectangle (5.0,3.800);
\draw[white,line width=0.3pt] (0.500,0) -- (0.500,3.800);
\draw[white,line width=0.3pt] (1.000,0) -- (1.000,3.800);
\draw[white,line width=0.3pt] (1.500,0) -- (1.500,3.800);
\draw[white,line width=0.3pt] (2.000,0) -- (2.000,3.800);
\draw[white,line width=0.3pt] (2.500,0) -- (2.500,3.800);
\draw[white,line width=0.3pt] (3.000,0) -- (3.000,3.800);
\draw[white,line width=0.3pt] (3.500,0) -- (3.500,3.800);
\draw[white,line width=0.3pt] (4.000,0) -- (4.000,3.800);
\draw[white,line width=0.3pt] (4.500,0) -- (4.500,3.800);
\draw[white,line width=0.3pt] (5.000,0) -- (5.000,3.800);
\draw[white,line width=0.3pt] (0,0.000) -- (5.0,0.000);
\node[anchor=east,text=vzInkTwo,inner sep=1.5pt] at (0,0.000) {36};
\draw[white,line width=0.3pt] (0,0.585) -- (5.0,0.585);
\node[anchor=east,text=vzInkTwo,inner sep=1.5pt] at (0,0.585) {38};
\draw[white,line width=0.3pt] (0,1.169) -- (5.0,1.169);
\node[anchor=east,text=vzInkTwo,inner sep=1.5pt] at (0,1.169) {40};
\draw[white,line width=0.3pt] (0,1.754) -- (5.0,1.754);
\node[anchor=east,text=vzInkTwo,inner sep=1.5pt] at (0,1.754) {42};
\draw[white,line width=0.3pt] (0,2.338) -- (5.0,2.338);
\node[anchor=east,text=vzInkTwo,inner sep=1.5pt] at (0,2.338) {44};
\draw[white,line width=0.3pt] (0,2.923) -- (5.0,2.923);
\node[anchor=east,text=vzInkTwo,inner sep=1.5pt] at (0,2.923) {46};
\draw[white,line width=0.3pt] (0,3.508) -- (5.0,3.508);
\node[anchor=east,text=vzInkTwo,inner sep=1.5pt] at (0,3.508) {48};
\draw[vzAxis,line width=0.4pt] (0,0) -- (5.0,0);
\draw[vzAxis,line width=0.4pt] (0.500,0) -- (0.500,-0.06);
\node[anchor=north,text=vzInkTwo,inner sep=1.5pt] at (0.500,-0.06) {0.1};
\draw[vzAxis,line width=0.4pt] (1.000,0) -- (1.000,-0.06);
\node[anchor=north,text=vzInkTwo,inner sep=1.5pt] at (1.000,-0.06) {0.2};
\draw[vzAxis,line width=0.4pt] (1.500,0) -- (1.500,-0.06);
\node[anchor=north,text=vzInkTwo,inner sep=1.5pt] at (1.500,-0.06) {0.3};
\draw[vzAxis,line width=0.4pt] (2.000,0) -- (2.000,-0.06);
\node[anchor=north,text=vzInkTwo,inner sep=1.5pt] at (2.000,-0.06) {0.4};
\draw[vzAxis,line width=0.4pt] (2.500,0) -- (2.500,-0.06);
\node[anchor=north,text=vzInkTwo,inner sep=1.5pt] at (2.500,-0.06) {0.5};
\draw[vzAxis,line width=0.4pt] (3.000,0) -- (3.000,-0.06);
\node[anchor=north,text=vzInkTwo,inner sep=1.5pt] at (3.000,-0.06) {0.6};
\draw[vzAxis,line width=0.4pt] (3.500,0) -- (3.500,-0.06);
\node[anchor=north,text=vzInkTwo,inner sep=1.5pt] at (3.500,-0.06) {0.7};
\draw[vzAxis,line width=0.4pt] (4.000,0) -- (4.000,-0.06);
\node[anchor=north,text=vzInkTwo,inner sep=1.5pt] at (4.000,-0.06) {0.8};
\draw[vzAxis,line width=0.4pt] (4.500,0) -- (4.500,-0.06);
\node[anchor=north,text=vzInkTwo,inner sep=1.5pt] at (4.500,-0.06) {0.9};
\draw[vzAxis,line width=0.4pt] (5.000,0) -- (5.000,-0.06);
\node[anchor=north,text=vzInkTwo,inner sep=1.5pt] at (5.000,-0.06) {1};
\node[anchor=north,text=vzInk] at (2.500,-0.36) {Temperature $T$};
\node[anchor=south,rotate=90,text=vzInk] at (-0.42,1.900) {Exact solve rate (\%)};
\fill[vzBase,opacity=0.13] (0.500,1.174) -- (1.000,1.224) -- (1.500,1.247) -- (2.000,1.249) -- (2.500,1.252) -- (3.000,1.265) -- (4.000,1.304) -- (5.000,1.303) -- (5.000,2.030) -- (4.000,1.980) -- (3.000,1.912) -- (2.500,1.837) -- (2.000,1.820) -- (1.500,1.803) -- (1.000,1.796) -- (0.500,1.759) -- cycle;
\fill[vzT4,opacity=0.13] (0.500,1.997) -- (1.000,1.968) -- (1.500,1.922) -- (2.000,1.824) -- (2.500,1.795) -- (3.000,1.776) -- (4.000,1.575) -- (5.000,1.315) -- (5.000,2.027) -- (4.000,2.264) -- (3.000,2.375) -- (2.500,2.483) -- (2.000,2.541) -- (1.500,2.599) -- (1.000,2.689) -- (0.500,2.806) -- cycle;
\fill[vzT2,opacity=0.13] (0.500,2.108) -- (1.000,2.209) -- (1.500,2.003) -- (2.000,1.971) -- (2.500,1.791) -- (3.000,1.598) -- (4.000,1.561) -- (5.000,1.306) -- (5.000,2.036) -- (4.000,2.152) -- (3.000,2.300) -- (2.500,2.282) -- (2.000,2.268) -- (1.500,2.275) -- (1.000,2.254) -- (0.500,2.374) -- cycle;
\fill[vzT1,opacity=0.13] (0.500,2.802) -- (1.000,2.722) -- (1.500,2.586) -- (2.000,2.402) -- (2.500,2.271) -- (3.000,2.073) -- (4.000,1.743) -- (5.000,1.330) -- (5.000,1.993) -- (4.000,2.174) -- (3.000,2.770) -- (2.500,3.030) -- (2.000,3.181) -- (1.500,3.221) -- (1.000,3.349) -- (0.500,3.512) -- cycle;
\draw[vzBase,line width=1.1pt,line join=round,line cap=round,dash pattern=on 3pt off 1.6pt] (0.500,1.466) -- (1.000,1.510) -- (1.500,1.525) -- (2.000,1.535) -- (2.500,1.544) -- (3.000,1.588) -- (4.000,1.642) -- (5.000,1.666);
\draw[vzT4,line width=1.1pt,line join=round,line cap=round,dash pattern=on 3pt off 1.6pt] (0.500,2.402) -- (1.000,2.329) -- (1.500,2.261) -- (2.000,2.183) -- (2.500,2.139) -- (3.000,2.075) -- (4.000,1.919) -- (5.000,1.671);
\draw[vzT2,line width=1.1pt,line join=round,line cap=round,dash pattern=on 3pt off 1.6pt] (0.500,2.241) -- (1.000,2.231) -- (1.500,2.139) -- (2.000,2.119) -- (2.500,2.036) -- (3.000,1.949) -- (4.000,1.856) -- (5.000,1.671);
\draw[vzT1,line width=1.1pt,line join=round,line cap=round] (0.500,3.157) -- (1.000,3.035) -- (1.500,2.904) -- (2.000,2.792) -- (2.500,2.650) -- (3.000,2.421) -- (4.000,1.958) -- (5.000,1.661);
\filldraw[fill=white,draw=vzBase,line width=0.75pt] (0.500,1.466) circle (0.059);
\filldraw[fill=white,draw=vzBase,line width=0.75pt] (1.000,1.510) circle (0.059);
\filldraw[fill=white,draw=vzBase,line width=0.75pt] (1.500,1.525) circle (0.059);
\filldraw[fill=white,draw=vzBase,line width=0.75pt] (2.000,1.535) circle (0.059);
\filldraw[fill=white,draw=vzBase,line width=0.75pt] (2.500,1.544) circle (0.059);
\filldraw[fill=white,draw=vzBase,line width=0.75pt] (3.000,1.588) circle (0.059);
\filldraw[fill=white,draw=vzBase,line width=0.75pt] (4.000,1.642) circle (0.059);
\filldraw[fill=white,draw=vzBase,line width=0.75pt] (5.000,1.666) circle (0.059);
\filldraw[fill=white,draw=vzT4,line width=0.75pt] (0.449,2.350) rectangle (0.551,2.453);
\filldraw[fill=white,draw=vzT4,line width=0.75pt] (0.949,2.277) rectangle (1.051,2.380);
\filldraw[fill=white,draw=vzT4,line width=0.75pt] (1.449,2.209) rectangle (1.551,2.312);
\filldraw[fill=white,draw=vzT4,line width=0.75pt] (1.949,2.131) rectangle (2.051,2.234);
\filldraw[fill=white,draw=vzT4,line width=0.75pt] (2.449,2.087) rectangle (2.551,2.190);
\filldraw[fill=white,draw=vzT4,line width=0.75pt] (2.949,2.024) rectangle (3.051,2.127);
\filldraw[fill=white,draw=vzT4,line width=0.75pt] (3.949,1.868) rectangle (4.051,1.971);
\filldraw[fill=white,draw=vzT4,line width=0.75pt] (4.949,1.620) rectangle (5.051,1.723);
\filldraw[fill=white,draw=vzT2,line width=0.75pt] (0.500,2.314) -- (0.437,2.204) -- (0.563,2.204) -- cycle;
\filldraw[fill=white,draw=vzT2,line width=0.75pt] (1.000,2.304) -- (0.937,2.195) -- (1.063,2.195) -- cycle;
\filldraw[fill=white,draw=vzT2,line width=0.75pt] (1.500,2.212) -- (1.437,2.102) -- (1.563,2.102) -- cycle;
\filldraw[fill=white,draw=vzT2,line width=0.75pt] (2.000,2.192) -- (1.937,2.083) -- (2.063,2.083) -- cycle;
\filldraw[fill=white,draw=vzT2,line width=0.75pt] (2.500,2.110) -- (2.437,2.000) -- (2.563,2.000) -- cycle;
\filldraw[fill=white,draw=vzT2,line width=0.75pt] (3.000,2.022) -- (2.937,1.912) -- (3.063,1.912) -- cycle;
\filldraw[fill=white,draw=vzT2,line width=0.75pt] (4.000,1.929) -- (3.937,1.820) -- (4.063,1.820) -- cycle;
\filldraw[fill=white,draw=vzT2,line width=0.75pt] (5.000,1.744) -- (4.937,1.634) -- (5.063,1.634) -- cycle;
\filldraw[fill=vzT1,draw=white,line width=0.5pt] (0.500,3.238) -- (0.581,3.157) -- (0.500,3.076) -- (0.419,3.157) -- cycle;
\filldraw[fill=vzT1,draw=white,line width=0.5pt] (1.000,3.116) -- (1.081,3.035) -- (1.000,2.954) -- (0.919,3.035) -- cycle;
\filldraw[fill=vzT1,draw=white,line width=0.5pt] (1.500,2.985) -- (1.581,2.904) -- (1.500,2.822) -- (1.419,2.904) -- cycle;
\filldraw[fill=vzT1,draw=white,line width=0.5pt] (2.000,2.873) -- (2.081,2.792) -- (2.000,2.710) -- (1.919,2.792) -- cycle;
\filldraw[fill=vzT1,draw=white,line width=0.5pt] (2.500,2.732) -- (2.581,2.650) -- (2.500,2.569) -- (2.419,2.650) -- cycle;
\filldraw[fill=vzT1,draw=white,line width=0.5pt] (3.000,2.503) -- (3.081,2.421) -- (3.000,2.340) -- (2.919,2.421) -- cycle;
\filldraw[fill=vzT1,draw=white,line width=0.5pt] (4.000,2.040) -- (4.081,1.958) -- (4.000,1.877) -- (3.919,1.958) -- cycle;
\filldraw[fill=vzT1,draw=white,line width=0.5pt] (5.000,1.743) -- (5.081,1.661) -- (5.000,1.580) -- (4.919,1.661) -- cycle;
\draw[vzBase,line width=1.1pt,dash pattern=on 3pt off 1.6pt] (0.150,0.430) -- (0.570,0.430);
\filldraw[fill=white,draw=vzBase,line width=0.75pt] (0.360,0.430) circle (0.059);
\node[anchor=west,text=vzInk,inner sep=1pt] at (0.650,0.430) {Cont., standard};
\draw[vzT4,line width=1.1pt,dash pattern=on 3pt off 1.6pt] (2.750,0.430) -- (3.170,0.430);
\filldraw[fill=white,draw=vzT4,line width=0.75pt] (2.909,0.379) rectangle (3.011,0.481);
\node[anchor=west,text=vzInk,inner sep=1pt] at (3.250,0.430) {Cont., BCT-E};
\draw[vzT2,line width=1.1pt,dash pattern=on 3pt off 1.6pt] (0.150,0.120) -- (0.570,0.120);
\filldraw[fill=white,draw=vzT2,line width=0.75pt] (0.360,0.193) -- (0.297,0.083) -- (0.423,0.083) -- cycle;
\node[anchor=west,text=vzInk,inner sep=1pt] at (0.650,0.120) {Cont., BCT-P};
\draw[vzT1,line width=1.1pt] (2.750,0.120) -- (3.170,0.120);
\filldraw[fill=vzT1,draw=white,line width=0.5pt] (2.960,0.201) -- (3.041,0.120) -- (2.960,0.039) -- (2.879,0.120) -- cycle;
\node[anchor=west,text=vzInk,inner sep=1pt] at (3.250,0.120) {TFT, BCT-L};
\end{tikzpicture}
\caption{Sudoku-Hard exact solve rate (NFE $=180$): mean and $\pm$ one sample standard deviation over three generation seeds; base-checkpoint results are in Table~\ref{tab:app-sudoku}.}
\label{fig:sudoku}
\vspace{-10pt}
\end{wrapfigure}

Sudoku-Hard tests whether our conclusions carry over to tightly constrained generation, in which each puzzle has a single valid completion.
Because no Sudoku checkpoint is released, we train the base model from scratch and obtain TFT by continuing it; as in Section~\ref{sec:exp-main}, the fair baseline for BCT-L is therefore \sflmcont{} (Figure~\ref{fig:sudoku}).
Following the released configuration of \sflm{}~\citep{hypersphericalflows2026}, we sample with 180 NFE.

Lowering the temperature slightly hurts standard tempering, whereas BCT-L improves monotonically to 46.8\%.
The training-free variants also benefit from lower temperatures, and on this task BCT-P does not collapse; still, BCT-L is the most accurate at every $T\le0.6$, with a margin that widens as $T$ decreases, and the base checkpoint shows the same pattern (Appendix~\ref{app:sudoku}).


\subsection{Beyond the Geodesic Bridge: von Mises--Fisher Flows}
\label{sec:exp-sfcd}

\begin{wrapfigure}{r}{0.47\linewidth}
\vspace{-12pt}
\centering
\begin{tikzpicture}[x=1cm,y=1cm,font=\scriptsize]
\definecolor{vzBlue}{HTML}{2A78D6}
\definecolor{vzOrange}{HTML}{EB6834}
\definecolor{vzAqua}{HTML}{1BAF7A}
\definecolor{vzBase}{HTML}{6B6A66}
\definecolor{vzMuted}{HTML}{9A9892}
\definecolor{vzGrid}{HTML}{E1E0D9}
\definecolor{vzAxis}{HTML}{B5B4AC}
\definecolor{vzInk}{HTML}{0B0B0B}
\definecolor{vzInkTwo}{HTML}{52514E}
\definecolor{vzViolet}{HTML}{4A3AA7}
\definecolor{vzT4}{HTML}{86B6EF}
\definecolor{vzT2}{HTML}{2A78D6}
\definecolor{vzT1}{HTML}{104281}
\definecolor{vzBg}{HTML}{F2F2F2}
\fill[vzBg] (0,0) rectangle (5.0,3.800);
\draw[white,line width=0.3pt] (0.500,0) -- (0.500,3.800);
\draw[white,line width=0.3pt] (1.000,0) -- (1.000,3.800);
\draw[white,line width=0.3pt] (1.500,0) -- (1.500,3.800);
\draw[white,line width=0.3pt] (2.000,0) -- (2.000,3.800);
\draw[white,line width=0.3pt] (2.500,0) -- (2.500,3.800);
\draw[white,line width=0.3pt] (3.000,0) -- (3.000,3.800);
\draw[white,line width=0.3pt] (3.500,0) -- (3.500,3.800);
\draw[white,line width=0.3pt] (4.000,0) -- (4.000,3.800);
\draw[white,line width=0.3pt] (4.500,0) -- (4.500,3.800);
\draw[white,line width=0.3pt] (5.000,0) -- (5.000,3.800);
\draw[white,line width=0.3pt] (0,0.000) -- (5.0,0.000);
\node[anchor=east,text=vzInkTwo,inner sep=1.5pt] at (0,0.000) {20};
\draw[white,line width=0.3pt] (0,1.267) -- (5.0,1.267);
\node[anchor=east,text=vzInkTwo,inner sep=1.5pt] at (0,1.267) {25};
\draw[white,line width=0.3pt] (0,2.533) -- (5.0,2.533);
\node[anchor=east,text=vzInkTwo,inner sep=1.5pt] at (0,2.533) {30};
\draw[white,line width=0.3pt] (0,3.800) -- (5.0,3.800);
\node[anchor=east,text=vzInkTwo,inner sep=1.5pt] at (0,3.800) {35};
\draw[vzAxis,line width=0.4pt] (0,0) -- (5.0,0);
\draw[vzAxis,line width=0.4pt] (0.500,0) -- (0.500,-0.06);
\node[anchor=north,text=vzInkTwo,inner sep=1.5pt] at (0.500,-0.06) {0.1};
\draw[vzAxis,line width=0.4pt] (1.000,0) -- (1.000,-0.06);
\node[anchor=north,text=vzInkTwo,inner sep=1.5pt] at (1.000,-0.06) {0.2};
\draw[vzAxis,line width=0.4pt] (1.500,0) -- (1.500,-0.06);
\node[anchor=north,text=vzInkTwo,inner sep=1.5pt] at (1.500,-0.06) {0.3};
\draw[vzAxis,line width=0.4pt] (2.000,0) -- (2.000,-0.06);
\node[anchor=north,text=vzInkTwo,inner sep=1.5pt] at (2.000,-0.06) {0.4};
\draw[vzAxis,line width=0.4pt] (2.500,0) -- (2.500,-0.06);
\node[anchor=north,text=vzInkTwo,inner sep=1.5pt] at (2.500,-0.06) {0.5};
\draw[vzAxis,line width=0.4pt] (3.000,0) -- (3.000,-0.06);
\node[anchor=north,text=vzInkTwo,inner sep=1.5pt] at (3.000,-0.06) {0.6};
\draw[vzAxis,line width=0.4pt] (3.500,0) -- (3.500,-0.06);
\node[anchor=north,text=vzInkTwo,inner sep=1.5pt] at (3.500,-0.06) {0.7};
\draw[vzAxis,line width=0.4pt] (4.000,0) -- (4.000,-0.06);
\node[anchor=north,text=vzInkTwo,inner sep=1.5pt] at (4.000,-0.06) {0.8};
\draw[vzAxis,line width=0.4pt] (4.500,0) -- (4.500,-0.06);
\node[anchor=north,text=vzInkTwo,inner sep=1.5pt] at (4.500,-0.06) {0.9};
\draw[vzAxis,line width=0.4pt] (5.000,0) -- (5.000,-0.06);
\node[anchor=north,text=vzInkTwo,inner sep=1.5pt] at (5.000,-0.06) {1};
\node[anchor=north,text=vzInk] at (2.500,-0.36) {Temperature $T$};
\node[anchor=south,rotate=90,text=vzInk] at (-0.42,1.900) {GSM8K accuracy (\%)};
\begin{scope}\clip (0,0) rectangle (5.0,3.800);
\fill[vzBase,opacity=0.13] (0.500,2.744) -- (1.000,2.980) -- (1.500,3.041) -- (2.000,2.973) -- (2.500,2.793) -- (3.000,2.493) -- (4.000,1.956) -- (5.000,0.732) -- (5.000,1.089) -- (4.000,2.093) -- (3.000,2.884) -- (2.500,3.184) -- (2.000,3.396) -- (1.500,3.351) -- (1.000,3.182) -- (0.500,2.957) -- cycle;
\fill[vzT4,opacity=0.13] (0.500,-5.067) -- (1.000,-5.071) -- (1.500,-4.388) -- (2.000,1.533) -- (2.500,2.720) -- (3.000,2.726) -- (4.000,2.291) -- (5.000,0.732) -- (5.000,1.089) -- (4.000,2.672) -- (3.000,3.121) -- (2.500,2.958) -- (2.000,1.694) -- (1.500,-4.262) -- (1.000,-5.054) -- (0.500,-5.067) -- cycle;
\fill[vzT2,opacity=0.13] (0.500,2.014) -- (1.000,3.029) -- (1.500,3.241) -- (2.000,3.383) -- (2.500,3.267) -- (3.000,3.036) -- (4.000,2.163) -- (5.000,0.732) -- (5.000,1.089) -- (4.000,2.539) -- (3.000,3.264) -- (2.500,3.463) -- (2.000,3.677) -- (1.500,3.619) -- (1.000,3.171) -- (0.500,2.288) -- cycle;
\fill[vzT1,opacity=0.13] (0.500,2.822) -- (1.000,3.070) -- (1.500,3.526) -- (2.000,3.385) -- (2.500,3.270) -- (3.000,2.934) -- (4.000,1.829) -- (5.000,0.175) -- (5.000,0.608) -- (4.000,2.320) -- (3.000,3.173) -- (2.500,3.590) -- (2.000,3.875) -- (1.500,3.803) -- (1.000,3.637) -- (0.500,3.093) -- cycle;
\draw[vzBase,line width=1.1pt,line join=round,line cap=round,dash pattern=on 3pt off 1.6pt] (0.500,2.850) -- (1.000,3.081) -- (1.500,3.196) -- (2.000,3.184) -- (2.500,2.989) -- (3.000,2.689) -- (4.000,2.024) -- (5.000,0.910);
\draw[vzT4,line width=1.1pt,line join=round,line cap=round,dash pattern=on 3pt off 1.6pt] (0.500,-5.067) -- (1.000,-5.063) -- (1.500,-4.325) -- (2.000,1.613) -- (2.500,2.839) -- (3.000,2.923) -- (4.000,2.481) -- (5.000,0.910);
\draw[vzT2,line width=1.1pt,line join=round,line cap=round,dash pattern=on 3pt off 1.6pt] (0.500,2.151) -- (1.000,3.100) -- (1.500,3.430) -- (2.000,3.530) -- (2.500,3.365) -- (3.000,3.150) -- (4.000,2.351) -- (5.000,0.910);
\draw[vzT1,line width=1.1pt,line join=round,line cap=round] (0.500,2.958) -- (1.000,3.353) -- (1.500,3.665) -- (2.000,3.630) -- (2.500,3.430) -- (3.000,3.054) -- (4.000,2.074) -- (5.000,0.392);
\filldraw[fill=white,draw=vzBase,line width=0.75pt] (0.500,2.850) circle (0.059);
\filldraw[fill=white,draw=vzBase,line width=0.75pt] (1.000,3.081) circle (0.059);
\filldraw[fill=white,draw=vzBase,line width=0.75pt] (1.500,3.196) circle (0.059);
\filldraw[fill=white,draw=vzBase,line width=0.75pt] (2.000,3.184) circle (0.059);
\filldraw[fill=white,draw=vzBase,line width=0.75pt] (2.500,2.989) circle (0.059);
\filldraw[fill=white,draw=vzBase,line width=0.75pt] (3.000,2.689) circle (0.059);
\filldraw[fill=white,draw=vzBase,line width=0.75pt] (4.000,2.024) circle (0.059);
\filldraw[fill=white,draw=vzBase,line width=0.75pt] (5.000,0.910) circle (0.059);
\filldraw[fill=white,draw=vzT4,line width=0.75pt] (0.449,-5.118) rectangle (0.551,-5.015);
\filldraw[fill=white,draw=vzT4,line width=0.75pt] (0.949,-5.114) rectangle (1.051,-5.011);
\filldraw[fill=white,draw=vzT4,line width=0.75pt] (1.449,-4.377) rectangle (1.551,-4.274);
\filldraw[fill=white,draw=vzT4,line width=0.75pt] (1.949,1.562) rectangle (2.051,1.665);
\filldraw[fill=white,draw=vzT4,line width=0.75pt] (2.449,2.787) rectangle (2.551,2.890);
\filldraw[fill=white,draw=vzT4,line width=0.75pt] (2.949,2.872) rectangle (3.051,2.975);
\filldraw[fill=white,draw=vzT4,line width=0.75pt] (3.949,2.430) rectangle (4.051,2.533);
\filldraw[fill=white,draw=vzT4,line width=0.75pt] (4.949,0.859) rectangle (5.051,0.962);
\filldraw[fill=white,draw=vzT2,line width=0.75pt] (0.500,2.224) -- (0.437,2.115) -- (0.563,2.115) -- cycle;
\filldraw[fill=white,draw=vzT2,line width=0.75pt] (1.000,3.173) -- (0.937,3.063) -- (1.063,3.063) -- cycle;
\filldraw[fill=white,draw=vzT2,line width=0.75pt] (1.500,3.503) -- (1.437,3.394) -- (1.563,3.394) -- cycle;
\filldraw[fill=white,draw=vzT2,line width=0.75pt] (2.000,3.603) -- (1.937,3.494) -- (2.063,3.494) -- cycle;
\filldraw[fill=white,draw=vzT2,line width=0.75pt] (2.500,3.438) -- (2.437,3.328) -- (2.563,3.328) -- cycle;
\filldraw[fill=white,draw=vzT2,line width=0.75pt] (3.000,3.223) -- (2.937,3.113) -- (3.063,3.113) -- cycle;
\filldraw[fill=white,draw=vzT2,line width=0.75pt] (4.000,2.424) -- (3.937,2.314) -- (4.063,2.314) -- cycle;
\filldraw[fill=white,draw=vzT2,line width=0.75pt] (5.000,0.984) -- (4.937,0.874) -- (5.063,0.874) -- cycle;
\filldraw[fill=vzT1,draw=white,line width=0.5pt] (0.500,3.039) -- (0.581,2.958) -- (0.500,2.877) -- (0.419,2.958) -- cycle;
\filldraw[fill=vzT1,draw=white,line width=0.5pt] (1.000,3.435) -- (1.081,3.353) -- (1.000,3.272) -- (0.919,3.353) -- cycle;
\filldraw[fill=vzT1,draw=white,line width=0.5pt] (1.500,3.746) -- (1.581,3.665) -- (1.500,3.583) -- (1.419,3.665) -- cycle;
\filldraw[fill=vzT1,draw=white,line width=0.5pt] (2.000,3.711) -- (2.081,3.630) -- (2.000,3.549) -- (1.919,3.630) -- cycle;
\filldraw[fill=vzT1,draw=white,line width=0.5pt] (2.500,3.512) -- (2.581,3.430) -- (2.500,3.349) -- (2.419,3.430) -- cycle;
\filldraw[fill=vzT1,draw=white,line width=0.5pt] (3.000,3.135) -- (3.081,3.054) -- (3.000,2.973) -- (2.919,3.054) -- cycle;
\filldraw[fill=vzT1,draw=white,line width=0.5pt] (4.000,2.156) -- (4.081,2.074) -- (4.000,1.993) -- (3.919,2.074) -- cycle;
\filldraw[fill=vzT1,draw=white,line width=0.5pt] (5.000,0.473) -- (5.081,0.392) -- (5.000,0.311) -- (4.919,0.392) -- cycle;
\end{scope}
\draw[vzT1,line width=1.1pt] (2.100,0.120) -- (2.520,0.120);
\filldraw[fill=vzT1,draw=white,line width=0.5pt] (2.310,0.201) -- (2.391,0.120) -- (2.310,0.039) -- (2.229,0.120) -- cycle;
\node[anchor=west,text=vzInk,inner sep=1pt] at (2.600,0.120) {TFT, BCT-L};
\draw[vzT2,line width=1.1pt,dash pattern=on 3pt off 1.6pt] (2.100,0.430) -- (2.520,0.430);
\filldraw[fill=white,draw=vzT2,line width=0.75pt] (2.310,0.503) -- (2.247,0.393) -- (2.373,0.393) -- cycle;
\node[anchor=west,text=vzInk,inner sep=1pt] at (2.600,0.430) {Cont., BCT-P};
\draw[vzT4,line width=1.1pt,dash pattern=on 3pt off 1.6pt] (2.100,0.740) -- (2.520,0.740);
\filldraw[fill=white,draw=vzT4,line width=0.75pt] (2.259,0.689) rectangle (2.361,0.791);
\node[anchor=west,text=vzInk,inner sep=1pt] at (2.600,0.740) {Cont., BCT-E};
\draw[vzBase,line width=1.1pt,dash pattern=on 3pt off 1.6pt] (2.100,1.050) -- (2.520,1.050);
\filldraw[fill=white,draw=vzBase,line width=0.75pt] (2.310,1.050) circle (0.059);
\node[anchor=west,text=vzInk,inner sep=1pt] at (2.600,1.050) {Cont., standard};
\end{tikzpicture}
\caption{GSM8K accuracy of BCT on the vMF flow LM (NFE $=1024$): mean and $\pm$ one sample standard deviation over five generation seeds. BCT-E falls below the plotted range for $T\le0.3$; more detailed results are in Table~\ref{tab:app-sfcd}.}
\label{fig:sfcd}
\vspace{-10pt}
\end{wrapfigure}
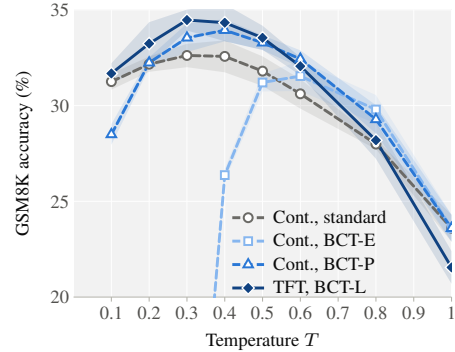

We finally apply BCT to SFCD~\citep{sphericalflows2026}, whose von Mises--Fisher bridge replaces the geodesic bridge of \sflm{} (Section~\ref{sec:bct-instantiations}).
We use its released TinyGSM checkpoint and sampler with NFE $=1024$, and updates for TFT and \sflmcont{}, and average over five generation seeds (Appendix~\ref{app:sfcd}).
Unlike \sflm{}, this flow already benefits from standard tempering: lowering the temperature raises its accuracy by 9 points between $T=1$ and $T=0.3$ (Figure~\ref{fig:sfcd}), suggesting that inconsistent tempering harms the vMF bridge less.
We therefore ask whether consistent tempering still improves a model whose standard tempering is already effective.

It does, mainly at moderate temperatures.
BCT-L improves on standard tempering at every $T\in[0.2,0.6]$ and attains the highest accuracy among the evaluated samplers, $34.5\pm0.5\%$ at $T=0.3$, against $32.6\pm0.6\%$ for standard tempering, a margin well beyond the seed standard deviations.

Under strong tempering, the learned realization remains stable while the training-free variants fail.
BCT-L declines at the lowest temperature, a trend absent on \sflm{}, but only to the level of standard tempering (31.7\% vs.\ 31.3\% at $T=0.1$).
Both training-free variants instead fall below it: BCT-P as on \sflm{} (28.5\%), and BCT-E, which is stable on \sflm{}, collapses to essentially zero accuracy for $T\le0.2$, likely because on this bridge its correction $-(\beta-1)\kappa_t\langle\ve_x,\vz\rangle$ amplifies any mismatch between the learned logits and the analytic bridge term by $\beta$ and $\kappa_t$.
BCT-L is thus the only variant that never falls below standard tempering for $T<1$: consistent tempering improves even a flow that already responds well to temperature, and training on the tempered bridge keeps it reliable under strong tempering.

\section{Related Work}
\label{sec:related}

\textbf{Posterior-induced flow language models.}
Non-autoregressive language models include discrete diffusion \citep{austin2021structured,lou2024discrete,sahoo2024simple,shi2024simplified,sahoo2025the,arriolablock,nie2025largelanguagediffusionmodels} and continuous diffusion and flow models, which map tokens to continuous representations such as learned embeddings or one-hot vectors \citep{li2022diffusion,dieleman2022continuous,gulrajani2023likelihood,fmlm2026,elf2026}.
In many continuous models, a denoiser trained with cross-entropy predicts the clean token, and its posterior weights conditional velocities, drifts, or scores.
Among spherical constructions \citep{davis2024fisher,jo2025continuous}, \sflm{} moves latents along geodesics of the hypersphere \citep{hypersphericalflows2026,chen2024flow}, whereas SFCD uses von Mises--Fisher bridges \citep{sphericalflows2026}; we instantiate BCT for both.

\textbf{Temperature and consistency.}
In autoregressive models and discrete diffusion, temperature rescales the categorical distribution from which tokens are sampled \citep{ackley1985learning,holtzman2020curious}, and lowering it substantially improves discrete diffusion language models on GSM8K \citep{hypersphericalflows2026}; we reproduce this in Appendix~\ref{app:dlm-temp}.
Continuous diffusion and flow models instead temper the score or the token posterior \citep{dieleman2022continuous,gulrajani2023likelihood,hypersphericalflows2026,azangulov2026sampling}, and for continuous data tempering is known not to commute with noising: the scaled score is not the score of the noised tempered distribution \citep{du2023reduce}.
Work on this inconsistency targets other settings: Feynman--Kac correctors reweight and resample populations of particles along annealed marginals \citep{skreta2025feynman,hasan2026discrete}, \citet{li2026diversify} correct score scaling under Gaussian noising for flattening ($T>1$), and temporal score rescaling controls a local temperature that leaves the mixture weights unchanged \citep{xu2026temporal}.
None of them addresses flows induced by a token posterior over a finite vocabulary, where tempering the posterior implicitly changes the conditional bridge while leaving the conditional velocities untempered.
BCT corrects this through the bridge itself, along a single sampling trajectory without particles, and extends the correction to training.

\textbf{Train--inference mismatch.}
Models evaluated on states that differ from those seen in training suffer from exposure bias, both in autoregressive models \citep{ranzato2016sequence,bengio2015scheduled} and in diffusion models, where it is reduced by perturbing training inputs or shifting sampling time steps \citep{ning2023input,li2024alleviating}.
There, the mismatch arises from the model's own prediction errors.
Under tempering, it is instead introduced by a known change of the bridge, so TFT can draw its training states exactly from the tempered bridge that the sampler follows.

\section{Conclusion}
\label{sec:conclusion}

We introduced \textbf{Bayes-Consistent Tempering (BCT)}, which treats temperature as part of the generative flow rather than as a post-hoc transformation of the denoising posterior. BCT aligns the tempered endpoint distribution, intermediate-state distribution, training distribution, and sampling dynamics; its training-free variants recover parts of this consistency for conventionally trained checkpoints. Across hyperspherical flow LMs with two denoiser architectures and the von Mises--Fisher flow LM, BCT improves GSM8K generation over broad temperature ranges, while its learned realization remains robust under aggressive tempering where the post-hoc variants deteriorate. The gains also extend to Sudoku, suggesting that BCT's benefits extend to tightly constrained generation. Overall, our results indicate that temperature in posterior-induced flow language models is better treated as part of the flow itself, across both training and inference.

\section*{Limitations and Discussion}
BCT applies to posterior-induced flows for which the tempered bridge and its corresponding sampling dynamics can be derived or approximated; the concrete derivations in this work cover the hyperspherical bridge of \sflm{} and the von Mises--Fisher bridge of SFCD. Extending BCT to other bridge families is an important direction for future work.

\section*{Acknowledgments}
This research was conducted by AIST and Institute of Science Tokyo under FRONTia, a Japanese national program led by the Ministry of Economy, Trade and Industry of Japan (METI) and the New Energy and Industrial Technology Development Organization (NEDO), aimed at developing a domestic multimodal foundation model for AI robots and physical AI. This work was also  supported in part by JSPS KAKENHI Grant Number 25H01137 and JST K Program Japan Grant Number JPMJKP24C3.

\bibliography{iclr2027_conference}

@misc{dit,
      title={Scalable Diffusion Models with Transformers}, 
      author={William Peebles and Saining Xie},
      year={2023},
      eprint={2212.09748},
      archivePrefix={arXiv},
      primaryClass={cs.CV},
      url={https://arxiv.org/abs/2212.09748}, 
}

@inproceedings{sudokukim,
  title={Fine-Tuning Masked Diffusion for Provable Self-Correction},
  author={Kim, Jaeyeon and Kim, Seunggeun and Lee, Taekyun and Pan, David Z. and Kim, Hyeji and Kakade, Sham and Chen, Sitan},
  booktitle={International Conference on Machine Learning},
  year={2026}
}

@misc{sudokuben,
      title={Accelerated Sampling from Masked Diffusion Models via Entropy Bounded Unmasking}, 
      author={Heli Ben-Hamu and Itai Gat and Daniel Severo and Niklas Nolte and Brian Karrer},
      year={2025},
      eprint={2505.24857},
      archivePrefix={arXiv},
      primaryClass={cs.LG},
      url={https://arxiv.org/abs/2505.24857}, 
}

@article{nie2025largelanguagediffusionmodels,
  title={Large language diffusion models},
  author={Nie, Shen and Zhu, Fengqi and You, Zebin and Zhang, Xiaolu and Ou, Jingyang and Hu, Jun and Zhou, Jun and Lin, Yankai and Wen, Ji-Rong and Li, Chongxuan},
  journal={arXiv preprint arXiv:2502.09992},
  year={2025}
}

@inproceedings{stark2024dirichlet,
  title={Dirichlet flow matching with applications to DNA sequence design},
  author={Stark, Hannes and Jing, Bowen and Wang, Chenyu and Corso, Gabriele and Berger, Bonnie and Barzilay, Regina and Jaakkola, Tommi},
  booktitle={Proceedings of the 41st International Conference on Machine Learning},
  pages={46495--46513},
  year={2024}
}

@inproceedings{
sahoo2025the,
title={The Diffusion Duality},
author={Subham Sekhar Sahoo and Justin Deschenaux and Aaron Gokaslan and Guanghan Wang and Justin T Chiu and Volodymyr Kuleshov},
booktitle={Forty-second International Conference on Machine Learning},
year={2025},
url={https://openreview.net/forum?id=9P9Y8FOSOk}
}

@inproceedings{arriolablock,
  title={Block Diffusion: Interpolating Between Autoregressive and Diffusion Language Models},
  author={Arriola, Marianne and Gokaslan, Aaron and Chiu, Justin T and Yang, Zhihan and Qi, Zhixuan and Han, Jiaqi and Sahoo, Subham Sekhar and Kuleshov, Volodymyr},
  booktitle={The Thirteenth International Conference on Learning Representations},
  year={2025}
}

@article{sahoo2024simple,
  title={Simple and effective masked diffusion language models},
  author={Sahoo, Subham and Arriola, Marianne and Schiff, Yair and Gokaslan, Aaron and Marroquin, Edgar and Chiu, Justin and Rush, Alexander and Kuleshov, Volodymyr},
  journal={Advances in Neural Information Processing Systems},
  volume={37},
  pages={130136--130184},
  year={2024}
}

@article{shi2024simplified,
  title={Simplified and generalized masked diffusion for discrete data},
  author={Shi, Jiaxin and Han, Kehang and Wang, Zhe and Doucet, Arnaud and Titsias, Michalis},
  journal={Advances in neural information processing systems},
  volume={37},
  pages={103131--103167},
  year={2024}
}

@article{li2022diffusion,
  title={Diffusion-lm improves controllable text generation},
  author={Li, Xiang and Thickstun, John and Gulrajani, Ishaan and Liang, Percy S and Hashimoto, Tatsunori B},
  journal={Advances in neural information processing systems},
  volume={35},
  pages={4328--4343},
  year={2022}
}

@article{chen2021codex,
  title={Evaluating large language models trained on code},
  author={Chen, Mark and Tworek, Jerry and Jun, Heewoo and Yuan, Qiming and Pinto, Henrique Ponde De Oliveira and Kaplan, Jared and Edwards, Harri and Burda, Yuri and Joseph, Nicholas and Brockman, Greg and others},
  journal={arXiv preprint arXiv:2107.03374},
  year={2021}
}

@article{elf2026,
  title={ELF: Embedded Language Flows},
  author={Hu, Keya and Qiu, Linlu and Lu, Yiyang and Zhao, Hanhong and Li, Tianhong and Kim, Yoon and Andreas, Jacob and He, Kaiming},
  journal={arXiv preprint arXiv:2605.10938},
  year={2026}
}

@article{fmlm2026,
  title={Flow map language models: One-step language modeling via continuous denoising},
  author={Lee, Chanhyuk and Yoo, Jaehoon and Agarwal, Manan and Shah, Sheel and Huang, Jerry and Raghunathan, Aditi and Hong, Seunghoon and Boffi, Nicholas M and Kim, Jinwoo},
  journal={arXiv preprint arXiv:2602.16813},
  year={2026}
}

@article{hypersphericalflows2026,
  title={Language Modeling with Hyperspherical Flows},
  author={Deschenaux, Justin and Gulcehre, Caglar},
  journal={arXiv preprint arXiv:2605.11125},
  year={2026}
}

@article{sphericalflows2026,
  title={Spherical Flows for Sampling Categorical Data},
  author={Chemseddine, Jannis and Kornhardt, Gregor and Steidl, Gabriele},
  journal={arXiv preprint arXiv:2605.05629},
  year={2026}
}

@article{liu2023tinygsm,
  title={{TinyGSM}: achieving $>$80\% on {GSM8K} with small language models},
  author={Liu, Bingbin and Bubeck, Sebastien and Eldan, Ronen and Kulkarni, Janardhan and Li, Yuanzhi and Nguyen, Anh and Ward, Rachel and Zhang, Yi},
  journal={arXiv preprint arXiv:2312.09241},
  year={2023}
}

@article{cobbe2021training,
  title={Training verifiers to solve math word problems},
  author={Cobbe, Karl and Kosaraju, Vineet and Bavarian, Mohammad and Chen, Mark and Jun, Heewoo and Kaiser, Lukasz and Plappert, Matthias and Tworek, Jerry and Hilton, Jacob and Nakano, Reiichiro and others},
  journal={arXiv preprint arXiv:2110.14168},
  year={2021}
}

@inproceedings{chen2024flow,
  title={Flow Matching on General Geometries},
  author={Chen, Ricky T. Q. and Lipman, Yaron},
  booktitle={The Twelfth International Conference on Learning Representations},
  year={2024}
}

@article{ackley1985learning,
  title={A Learning Algorithm for {B}oltzmann Machines},
  author={Ackley, David H. and Hinton, Geoffrey E. and Sejnowski, Terrence J.},
  journal={Cognitive Science},
  volume={9},
  number={1},
  pages={147--169},
  year={1985}
}

@inproceedings{holtzman2020curious,
  title={The Curious Case of Neural Text Degeneration},
  author={Holtzman, Ari and Buys, Jan and Du, Li and Forbes, Maxwell and Choi, Yejin},
  booktitle={International Conference on Learning Representations},
  year={2020}
}

@inproceedings{skreta2025feynman,
  title={{F}eynman-{K}ac Correctors in Diffusion: Annealing, Guidance, and Product of Experts},
  author={Skreta, Marta and Akhound-Sadegh, Tara and Ohanesian, Viktor and Bondesan, Roberto and Aspuru-Guzik, Al{\'a}n and Doucet, Arnaud and Brekelmans, Rob and Tong, Alexander and Neklyudov, Kirill},
  booktitle={Proceedings of the 42nd International Conference on Machine Learning},
  series={Proceedings of Machine Learning Research},
  volume={267},
  pages={55906--55949},
  year={2025}
}

@article{hasan2026discrete,
  title={Discrete {F}eynman-{K}ac Correctors},
  author={Hasan, Mohsin and Ohanesian, Viktor and Gazizov, Artem and Bengio, Yoshua and Aspuru-Guzik, Al{\'a}n and Bondesan, Roberto and Skreta, Marta and Neklyudov, Kirill},
  journal={arXiv preprint arXiv:2601.10403},
  year={2026}
}

@inproceedings{xu2026temporal,
  title={Temporal Score Rescaling for Temperature Sampling in Diffusion and Flow Models},
  author={Xu, Yanbo and Wu, Yu and Park, Sungjae and Zhou, Zhizhuo and Tulsiani, Shubham},
  booktitle={International Conference on Machine Learning},
  year={2026}
}

@article{li2026diversify,
  title={Diversify Diffusion with Temperature Sampling and Variance-Corrective Time Shifting},
  author={Li, Peizhuo and Aksan, Emre and Ichim, Alexandru-Eugen and Beeler, Thabo and Sorkine-Hornung, Olga},
  journal={arXiv preprint arXiv:2607.10853},
  year={2026}
}

@inproceedings{du2023reduce,
  title={Reduce, Reuse, Recycle: Compositional Generation with Energy-Based Diffusion Models and {MCMC}},
  author={Du, Yilun and Durkan, Conor and Strudel, Robin and Tenenbaum, Joshua B. and Dieleman, Sander and Fergus, Rob and Sohl-Dickstein, Jascha and Doucet, Arnaud and Grathwohl, Will},
  booktitle={Proceedings of the 40th International Conference on Machine Learning},
  series={Proceedings of Machine Learning Research},
  volume={202},
  pages={8489--8510},
  year={2023}
}

@article{azangulov2026sampling,
  title={Sampling from Flow Language Models via Marginal-Conditioned Bridges},
  author={Azangulov, Iskander and Zhang, Leo},
  journal={arXiv preprint arXiv:2605.13681},
  year={2026}
}

@inproceedings{jo2025continuous,
  title={Continuous Diffusion Model for Language Modeling},
  author={Jo, Jaehyeong and Hwang, Sung Ju},
  booktitle={Advances in Neural Information Processing Systems},
  year={2025}
}

@inproceedings{davis2024fisher,
  title={Fisher Flow Matching for Generative Modeling over Discrete Data},
  author={Davis, Oscar and Kessler, Samuel and Petrache, Mircea and Ceylan, {\.I}smail {\.I}lkan and Bronstein, Michael and Bose, Avishek Joey},
  booktitle={Advances in Neural Information Processing Systems},
  year={2024}
}

@inproceedings{gulrajani2023likelihood,
  title={Likelihood-Based Diffusion Language Models},
  author={Gulrajani, Ishaan and Hashimoto, Tatsunori B.},
  booktitle={Advances in Neural Information Processing Systems},
  volume={36},
  year={2023}
}

@article{dieleman2022continuous,
  title={Continuous diffusion for categorical data},
  author={Dieleman, Sander and Sartran, Laurent and Roshannai, Arman and Savinov, Nikolay and Ganin, Yaroslav and Richemond, Pierre H. and Doucet, Arnaud and Strudel, Robin and Dyer, Chris and Durkan, Conor and Hawthorne, Curtis and Leblond, R{\'e}mi and Grathwohl, Will and Adler, Jonas},
  journal={arXiv preprint arXiv:2211.15089},
  year={2022}
}

@inproceedings{bengio2015scheduled,
  title={Scheduled Sampling for Sequence Prediction with Recurrent Neural Networks},
  author={Bengio, Samy and Vinyals, Oriol and Jaitly, Navdeep and Shazeer, Noam},
  booktitle={Advances in Neural Information Processing Systems},
  year={2015}
}

@inproceedings{ning2023input,
  title={Input Perturbation Reduces Exposure Bias in Diffusion Models},
  author={Ning, Mang and Sangineto, Enver and Porrello, Angelo and Calderara, Simone and Cucchiara, Rita},
  booktitle={International Conference on Machine Learning},
  year={2023}
}

@inproceedings{li2024alleviating,
  title={Alleviating Exposure Bias in Diffusion Models through Sampling with Shifted Time Steps},
  author={Li, Mingxiao and Qu, Tingyu and Yao, Ruicong and Sun, Wei and Moens, Marie-Francine},
  booktitle={International Conference on Learning Representations},
  year={2024}
}

@inproceedings{austin2021structured,
  title={Structured Denoising Diffusion Models in Discrete State-Spaces},
  author={Austin, Jacob and Johnson, Daniel D. and Ho, Jonathan and Tarlow, Daniel and van den Berg, Rianne},
  booktitle={Advances in Neural Information Processing Systems},
  year={2021}
}

@inproceedings{lou2024discrete,
  title={Discrete Diffusion Modeling by Estimating the Ratios of the Data Distribution},
  author={Lou, Aaron and Meng, Chenlin and Ermon, Stefano},
  booktitle={International Conference on Machine Learning},
  year={2024}
}

@inproceedings{ranzato2016sequence,
  title={Sequence Level Training with Recurrent Neural Networks},
  author={Ranzato, Marc'Aurelio and Chopra, Sumit and Auli, Michael and Zaremba, Wojciech},
  booktitle={International Conference on Learning Representations},
  year={2016}
}
\bibliographystyle{iclr2027_conference}

\clearpage
\appendix
\section{Sequences}
\label{app:sequences}

For a sequence $x=(x^1,\dots,x^L)$ with joint endpoint distribution $\pi(x)$ and a bridge that factorizes over positions, the posterior at position $l$ marginalizes the other positions:
\begin{equation}
r_t(x^l\mid\vz)
=\frac{q_t(\vz^l\mid x^l)\,\tilde\pi_t(x^l;\vz^{-l})}{\sum_{y}q_t(\vz^l\mid y)\,\tilde\pi_t(y;\vz^{-l})},
\qquad
\tilde\pi_t(x^l;\vz^{-l})=\sum_{x^{-l}}\pi(x)\prod_{j\neq l}q_t(\vz^j\mid x^j).
\label{eq:seq-post}
\end{equation}
This is \eqp{eq:bayes} at position $l$ with $\pi$ replaced by the context-conditioned endpoint $\tilde\pi_t(\cdot\,;\vz^{-l})$, which does not depend on $\vz^l$. 

Every construction of Section~\ref{sec:bct} therefore applies position-wise and tempers $\tilde\pi_t$; under TFT, $\tilde\pi_t$ is formed with the tempered bridges $q_{t,T}$ at the other positions. 
For any fixed endpoint sequence (x), the product of the position-wise conditional bridges is transported by the concatenation of their corresponding conditional velocities, so the joint conditional continuity equation holds by the product rule.
None of them tempers the joint endpoint: that would require the context factor $\sum_{x^{-l}}\pi(x)^{\beta}\prod_{j\neq l}q'_t(\vz^j\mid x^j)$, with $q'_t$ the bridge of the target pair, which in general is not proportional to the power of $\tilde\pi$ that every variant uses. Per-token temperature in autoregressive models has the same property.

\section{The hyperspherical bridge}
\label{app:sflm}

Throughout, $\theta=\theta_x(\vz)$ is the polar angle about $\ve_x$, $\nu=d-2$, $\sigma$ is the surface measure on $\sph$, and $a=a_t\in(0,1]$, which equals $1$ at the source and decreases in $t$; the bridge depends on $t$ only through $a$, and the results below hold for every $a\in(0,1]$.

\subsection{Bridge density}
\label{app:density}

\sflm{} draws $\vz_0$ uniformly on $\sph$ and sets $\vz_t=\mathrm{SLERP}(\vz_0,\ve_x;\alpha_t)$. About $\ve_x$, write $\vz=\cos\theta\,\ve_x+\sin\theta\,\bm\xi$ with $\bm\xi$ in the unit sphere $\mathbb{S}^{d-2}$ of the orthogonal complement of $\ve_x$. The surface measure factorizes as
\begin{equation}
d\sigma(\vz)=\sin^{\nu}\theta\;d\theta\;d\sigma_{d-2}(\bm\xi).
\label{eq:polar-measure}
\end{equation}
Let $\omega=\theta_x(\vz_0)$. Then $\mathrm{SLERP}(\vz_0,\ve_x;\alpha)=\bigl[\sin(a\omega)\,\vz_0+\sin((1-a)\omega)\,\ve_x\bigr]/\sin\omega$, and substituting $\vz_0=\cos\omega\,\ve_x+\sin\omega\,\bm\xi$ gives $\vz_t=\cos(a\omega)\,\ve_x+\sin(a\omega)\,\bm\xi$: SLERP keeps $\bm\xi$ and contracts the polar angle by the factor $a$. For uniform $\vz_0$, $(\omega,\bm\xi)$ has density $\sin^{\nu}\omega/|\sph|$ with respect to $d\omega\,d\sigma_{d-2}(\bm\xi)$. The map $\omega\mapsto a\omega$ has Jacobian $a$, and dividing by the factor $\sin^{\nu}\theta$ of \eqp{eq:polar-measure} gives
\begin{equation}
q_t(\vz\mid x)=\frac{\sin^{\nu}(\theta/a)}{a\,|\sph|\,\sin^{\nu}\theta}\;\ind\{\theta<a\pi\},
\label{eq:sflm-density}
\end{equation}
which is \eqp{eq:sflm-bridge}.

\subsection{Tempered bridge}
\label{app:tempered-bridge}

By \eqp{eq:polar-measure}, under $q_{t,T}(\cdot\mid x)\propto q_t(\cdot\mid x)^{\beta}$ the direction $\bm\xi$ is uniform on $\mathbb{S}^{d-2}$ and independent of the polar angle, whose marginal is
\begin{equation}
m_{a,T}(\theta)=\frac{g_{a,T}(\theta)}{C_{a,T}},
\qquad
g_{a,T}(\theta)=\sin^{\nu\beta}\!\Bigl(\frac{\theta}{a}\Bigr)\,\sin^{\nu(1-\beta)}\theta,
\qquad
\theta\in(0,a\pi),
\label{eq:polar}
\end{equation}
with $C_{a,T}=\int_0^{a\pi}g_{a,T}$. Rotations preserve $\sigma$ and map the bridges of different tokens onto one another, so $Z_{t,T}$ depends only on $(a,T,d)$. Since $g_{a,T}$ behaves like $\theta^{\nu}$ as $\theta\to0$ and like $(a\pi-\theta)^{\nu\beta}$ as $\theta\to a\pi$ ($a<1$), $Z_{t,T}<\infty$ and $q_{t,T}$ is a valid density for every $T>0$. Its support does not depend on $T$. At $a=1$, $g_{1,T}=\sin^{\nu}\theta$ for every $T$, so the tempered source $q_{0,T}$ is the uniform source of \sflm{}.

\subsection{Tempered conditional velocity}
\label{app:velocity}

We write the polar dynamics as $d\theta/da=v_{a,T}(\theta)$.

\textbf{Symmetry reduction.}
$q_{t,T}(\cdot\mid x)$ is invariant under the orthogonal transformations fixing $\ve_x$, and we take its conditional velocity to be equivariant, $u(R\vz)=Ru(\vz)$ for all such $R$. The transformations fixing both $\ve_x$ and $\vz$ therefore fix $u(\vz)$, and for $d\ge3$ the only tangent vectors at $\vz$ they fix are multiples of $\mathbf{e}_\theta$, the unit tangent pointing away from $\ve_x$ along the geodesic. Equivariance under transformations that move $\bm\xi$ makes the magnitude a function of $\theta$ alone, so $u=\dot\theta\,\mathbf{e}_\theta$.

\textbf{One-dimensional continuity equation.}
For a density $\rho$ with respect to $\sigma$, $\nabla\cdot(\rho\,v\,\mathbf{e}_\theta)=\sin^{-\nu}\theta\;\partial_\theta(\sin^{\nu}\theta\,\rho\,v)$. Multiplying $\partial_a\rho+\nabla\cdot(\rho\,v\,\mathbf{e}_\theta)=0$ by $|\mathbb{S}^{d-2}|\sin^{\nu}\theta$, which turns $\rho$ into the polar marginal $m_{a,T}$, gives
\begin{equation}
\partial_a m_{a,T}(\theta)+\partial_\theta\bigl(m_{a,T}(\theta)\,v_{a,T}(\theta)\bigr)=0 .
\label{eq:1d-continuity}
\end{equation}
With no flux through the pole ($m\,v\to0$ as $\theta\to0$), integrating over $(0,\theta)$ gives the unique solution $v_{a,T}=-\partial_aM_{a,T}/m_{a,T}$, where $M_{a,T}(\theta)=\int_0^{\theta}m_{a,T}$.

\textbf{Closed form.}
From \eqp{eq:polar}, $\partial_a g_{a,T}(\theta)=-(\nu\beta/a^{2})\,\theta\cot(\theta/a)\,g_{a,T}(\theta)$. Since $g_{a,T}(a\pi)=0$, differentiating $C_{a,T}$ produces no boundary term, and $\partial_aC_{a,T}=-(\nu\beta/a^{2})\,\bar H_{a,T}\,C_{a,T}$ with $\bar H_{a,T}=\E_{\theta\sim m_{a,T}}[\theta\cot(\theta/a)]$. Hence
\begin{equation}
v_{a,T}(\theta)=\frac{\nu\beta}{a^{2}}\,
\frac{\int_0^{\theta}\bigl[\vartheta\cot(\vartheta/a)-\bar H_{a,T}\bigr]\,g_{a,T}(\vartheta)\,d\vartheta}{g_{a,T}(\theta)}.
\label{eq:radial}
\end{equation}

\textbf{Velocity field.}
As $t$ increases, $da/dt=-\dot\alpha_t$, so $\dot\vz=-\dot\alpha_t\,v_{a,T}(\theta)\,\mathbf{e}_\theta$. Since $\Logm{\vz}(\ve_x)=-\theta\,\mathbf{e}_\theta$, this equals
\begin{equation*}
u_{t,T}(\vz\mid x)=w_{t,T}(\theta)\,u_t(\vz\mid x),
\qquad
w_{t,T}(\theta)=\frac{a_t\,v_{a_t,T}(\theta)}{\theta},
\qquad
u_t(\vz\mid x)=\frac{\dot\alpha_t}{a_t}\,\Logm{\vz}(\ve_x),
\end{equation*}
which is \eqp{eq:sflm-tempered-velocity}. At $T=1$, integration by parts with $s=\theta/a$ gives $\int_0^{\pi}s\cos s\,\sin^{\nu-1}s\,ds=-\frac1\nu\int_0^{\pi}\sin^{\nu}s\,ds$, hence $\bar H_{a,1}=-a/\nu$, and $\frac{d}{d\theta}\bigl[\frac{\theta}{a}g_{a,1}(\theta)\bigr]=\frac{\nu}{a^{2}}\bigl[\theta\cot(\theta/a)-\bar H_{a,1}\bigr]g_{a,1}(\theta)$. The integral in \eqp{eq:radial} thus equals $\frac{a}{\nu}\,\theta\,g_{a,1}(\theta)$, so $v_{a,1}=\theta/a$ and $w_{t,1}\equiv1$: $u_t$ is the \sflm{} conditional velocity, and the tempered construction reduces to it at $T=1$.

\subsection{Exact sampling of the tempered bridge}
\label{app:corruption}

Use the normalized angles $\psi=\omega/\pi$ and $\phi=\theta/(a\pi)$ in $(0,1)$. Let $F_{\mathrm{src}}$ be the CDF of the density $\propto\sin^{\nu}(\pi\psi)$, and let $F_{a,T}$ be the CDF of the density $\propto\sin^{\nu\beta}(\pi\phi)\sin^{\nu(1-\beta)}(a\pi\phi)$, which is \eqp{eq:polar} in the variable $\phi$. Define
\begin{equation}
Q_{a,T}(\omega)=a\pi\,F_{a,T}^{-1}\bigl(F_{\mathrm{src}}(\omega/\pi)\bigr).
\label{eq:transport}
\end{equation}
Under both the source and the tempered bridge, $\bm\xi$ is uniform on $\mathbb{S}^{d-2}$ and independent of the polar angle. Hence $\vz=\cos\theta\,\ve_x+\sin\theta\,\bm\xi$ with $\theta=Q_{a,T}(\omega)$ is an exact sample of $q_{t,T}(\cdot\mid x)$. At $\beta=1$, $F_{a,1}=F_{\mathrm{src}}$ and $\theta=a\omega$, which reproduces SLERP for the same $\vz_0$. Equivalently, $\vz=\mathrm{SLERP}(\vz_0,\ve_x;1-\theta/\omega)$, a SLERP with a position-specific parameter. For TFT we build the target CDF for each $(a_t,T)$ pair in the batch on a grid refined geometrically toward both ends of $[0,a_t\pi]$ and invert it at the drawn angles, with no interpolation table over $(a,T)$.

\section{Held-out selection of $\lambda$}
\label{app:practical}

\textbf{Held-out set.}
The TinyGSM data used by \sflm{} are split into a training part, which the TFT continuation reads, and a validation part of 118{,}216 problems, which it does not. From a random subset of 768 validation problems, we remove 111 prompts that also occur verbatim in the training part and 30 problems without a reliably recoverable reference answer (seven fall in both groups), leaving 634 problems, none of which matches a GSM8K test problem. The reference answer of each problem is the value returned by its TinyGSM reference program, and generated programs are graded as on GSM8K.

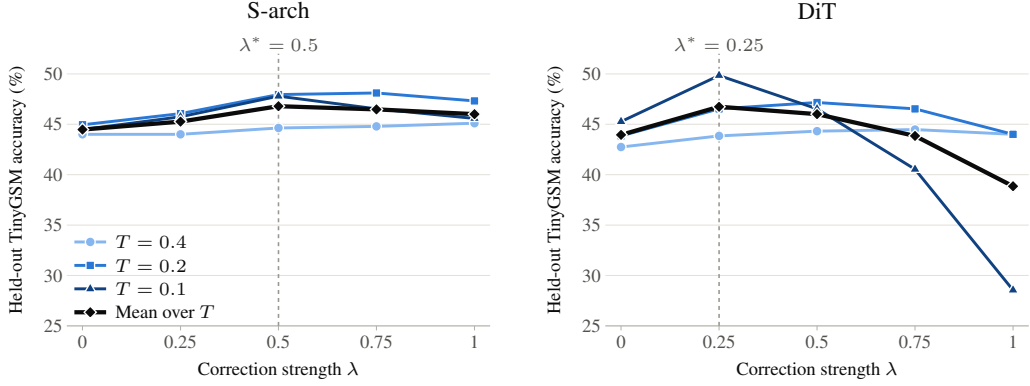
\begin{figure}[ht]
\centering
\begin{minipage}[t]{0.49\linewidth}
\centering
\begin{tikzpicture}[x=1cm,y=1cm,font=\scriptsize]
\definecolor{vzBlue}{HTML}{2A78D6}
\definecolor{vzOrange}{HTML}{EB6834}
\definecolor{vzAqua}{HTML}{1BAF7A}
\definecolor{vzBase}{HTML}{6B6A66}
\definecolor{vzMuted}{HTML}{9A9892}
\definecolor{vzGrid}{HTML}{E1E0D9}
\definecolor{vzAxis}{HTML}{B5B4AC}
\definecolor{vzInk}{HTML}{0B0B0B}
\definecolor{vzInkTwo}{HTML}{52514E}
\definecolor{vzViolet}{HTML}{4A3AA7}
\definecolor{vzT4}{HTML}{86B6EF}
\definecolor{vzT2}{HTML}{2A78D6}
\definecolor{vzT1}{HTML}{104281}
\draw[vzGrid,line width=0.3pt] (0,0.000) -- (5.6,0.000);
\node[anchor=east,text=vzInkTwo,inner sep=1.5pt] at (0,0.000) {25};
\draw[vzGrid,line width=0.3pt] (0,0.667) -- (5.6,0.667);
\node[anchor=east,text=vzInkTwo,inner sep=1.5pt] at (0,0.667) {30};
\draw[vzGrid,line width=0.3pt] (0,1.333) -- (5.6,1.333);
\node[anchor=east,text=vzInkTwo,inner sep=1.5pt] at (0,1.333) {35};
\draw[vzGrid,line width=0.3pt] (0,2.000) -- (5.6,2.000);
\node[anchor=east,text=vzInkTwo,inner sep=1.5pt] at (0,2.000) {40};
\draw[vzGrid,line width=0.3pt] (0,2.667) -- (5.6,2.667);
\node[anchor=east,text=vzInkTwo,inner sep=1.5pt] at (0,2.667) {45};
\draw[vzGrid,line width=0.3pt] (0,3.333) -- (5.6,3.333);
\node[anchor=east,text=vzInkTwo,inner sep=1.5pt] at (0,3.333) {50};
\draw[vzAxis,line width=0.4pt] (0,0) -- (5.6,0);
\draw[vzAxis,line width=0.4pt] (0.207,0) -- (0.207,-0.06);
\node[anchor=north,text=vzInkTwo,inner sep=1.5pt] at (0.207,-0.06) {0};
\draw[vzAxis,line width=0.4pt] (1.504,0) -- (1.504,-0.06);
\node[anchor=north,text=vzInkTwo,inner sep=1.5pt] at (1.504,-0.06) {0.25};
\draw[vzAxis,line width=0.4pt] (2.800,0) -- (2.800,-0.06);
\node[anchor=north,text=vzInkTwo,inner sep=1.5pt] at (2.800,-0.06) {0.5};
\draw[vzAxis,line width=0.4pt] (4.096,0) -- (4.096,-0.06);
\node[anchor=north,text=vzInkTwo,inner sep=1.5pt] at (4.096,-0.06) {0.75};
\draw[vzAxis,line width=0.4pt] (5.393,0) -- (5.393,-0.06);
\node[anchor=north,text=vzInkTwo,inner sep=1.5pt] at (5.393,-0.06) {1};
\node[anchor=north,text=vzInk] at (2.800,-0.36) {Correction strength $\lambda$};
\node[anchor=south,rotate=90,text=vzInk] at (-0.42,1.800) {Held-out TinyGSM accuracy (\%)};
\node[anchor=south,text=vzInk,font=\small] at (2.800,3.940) {S-arch};
\draw[vzMuted,line width=0.6pt,dash pattern=on 1.5pt off 1.5pt] (2.800,0) -- (2.800,3.600);
\node[anchor=south,text=vzInkTwo,inner sep=1pt] at (2.800,3.600) {$\lambda^*=0.5$};
\draw[vzT4,line width=1.1pt,line join=round,line cap=round] (0.207,2.534) -- (1.504,2.534) -- (2.800,2.618) -- (4.096,2.639) -- (5.393,2.681);
\draw[vzT2,line width=1.1pt,line join=round,line cap=round] (0.207,2.660) -- (1.504,2.808) -- (2.800,3.060) -- (4.096,3.081) -- (5.393,2.976);
\draw[vzT1,line width=1.1pt,line join=round,line cap=round] (0.207,2.597) -- (1.504,2.766) -- (2.800,3.039) -- (4.096,2.871) -- (5.393,2.744);
\draw[vzInk,line width=1.6pt,line join=round,line cap=round] (0.207,2.597) -- (1.504,2.702) -- (2.800,2.906) -- (4.096,2.864) -- (5.393,2.801);
\filldraw[fill=vzT4,draw=white,line width=0.5pt] (0.207,2.534) circle (0.065);
\filldraw[fill=vzT4,draw=white,line width=0.5pt] (1.504,2.534) circle (0.065);
\filldraw[fill=vzT4,draw=white,line width=0.5pt] (2.800,2.618) circle (0.065);
\filldraw[fill=vzT4,draw=white,line width=0.5pt] (4.096,2.639) circle (0.065);
\filldraw[fill=vzT4,draw=white,line width=0.5pt] (5.393,2.681) circle (0.065);
\filldraw[fill=vzT2,draw=white,line width=0.5pt] (0.150,2.603) rectangle (0.265,2.718);
\filldraw[fill=vzT2,draw=white,line width=0.5pt] (1.447,2.750) rectangle (1.561,2.865);
\filldraw[fill=vzT2,draw=white,line width=0.5pt] (2.743,3.003) rectangle (2.857,3.117);
\filldraw[fill=vzT2,draw=white,line width=0.5pt] (4.039,3.024) rectangle (4.153,3.138);
\filldraw[fill=vzT2,draw=white,line width=0.5pt] (5.335,2.919) rectangle (5.450,3.033);
\filldraw[fill=vzT1,draw=white,line width=0.5pt] (0.207,2.679) -- (0.137,2.557) -- (0.278,2.557) -- cycle;
\filldraw[fill=vzT1,draw=white,line width=0.5pt] (1.504,2.847) -- (1.433,2.725) -- (1.574,2.725) -- cycle;
\filldraw[fill=vzT1,draw=white,line width=0.5pt] (2.800,3.120) -- (2.730,2.998) -- (2.870,2.998) -- cycle;
\filldraw[fill=vzT1,draw=white,line width=0.5pt] (4.096,2.952) -- (4.026,2.830) -- (4.167,2.830) -- cycle;
\filldraw[fill=vzT1,draw=white,line width=0.5pt] (5.393,2.826) -- (5.322,2.704) -- (5.463,2.704) -- cycle;
\filldraw[fill=vzInk,draw=white,line width=0.5pt] (0.207,2.679) -- (0.289,2.597) -- (0.207,2.516) -- (0.126,2.597) -- cycle;
\filldraw[fill=vzInk,draw=white,line width=0.5pt] (1.504,2.784) -- (1.585,2.702) -- (1.504,2.621) -- (1.422,2.702) -- cycle;
\filldraw[fill=vzInk,draw=white,line width=0.5pt] (2.800,2.987) -- (2.881,2.906) -- (2.800,2.824) -- (2.719,2.906) -- cycle;
\filldraw[fill=vzInk,draw=white,line width=0.5pt] (4.096,2.945) -- (4.178,2.864) -- (4.096,2.782) -- (4.015,2.864) -- cycle;
\filldraw[fill=vzInk,draw=white,line width=0.5pt] (5.393,2.882) -- (5.474,2.801) -- (5.393,2.719) -- (5.311,2.801) -- cycle;
\draw[vzInk,line width=1.6pt] (0.100,0.180) -- (0.520,0.180);
\filldraw[fill=vzInk,draw=white,line width=0.5pt] (0.310,0.261) -- (0.391,0.180) -- (0.310,0.099) -- (0.229,0.180) -- cycle;
\node[anchor=west,text=vzInk,inner sep=1pt] at (0.600,0.180) {Mean over $T$};
\draw[vzT1,line width=1.1pt] (0.100,0.490) -- (0.520,0.490);
\filldraw[fill=vzT1,draw=white,line width=0.5pt] (0.310,0.571) -- (0.240,0.449) -- (0.380,0.449) -- cycle;
\node[anchor=west,text=vzInk,inner sep=1pt] at (0.600,0.490) {$T=0.1$};
\draw[vzT2,line width=1.1pt] (0.100,0.800) -- (0.520,0.800);
\filldraw[fill=vzT2,draw=white,line width=0.5pt] (0.253,0.743) rectangle (0.367,0.857);
\node[anchor=west,text=vzInk,inner sep=1pt] at (0.600,0.800) {$T=0.2$};
\draw[vzT4,line width=1.1pt] (0.100,1.110) -- (0.520,1.110);
\filldraw[fill=vzT4,draw=white,line width=0.5pt] (0.310,1.110) circle (0.065);
\node[anchor=west,text=vzInk,inner sep=1pt] at (0.600,1.110) {$T=0.4$};
\end{tikzpicture}
\end{minipage}\hfill
\begin{minipage}[t]{0.49\linewidth}
\centering
\begin{tikzpicture}[x=1cm,y=1cm,font=\scriptsize]
\definecolor{vzBlue}{HTML}{2A78D6}
\definecolor{vzOrange}{HTML}{EB6834}
\definecolor{vzAqua}{HTML}{1BAF7A}
\definecolor{vzBase}{HTML}{6B6A66}
\definecolor{vzMuted}{HTML}{9A9892}
\definecolor{vzGrid}{HTML}{E1E0D9}
\definecolor{vzAxis}{HTML}{B5B4AC}
\definecolor{vzInk}{HTML}{0B0B0B}
\definecolor{vzInkTwo}{HTML}{52514E}
\definecolor{vzViolet}{HTML}{4A3AA7}
\definecolor{vzT4}{HTML}{86B6EF}
\definecolor{vzT2}{HTML}{2A78D6}
\definecolor{vzT1}{HTML}{104281}
\draw[vzGrid,line width=0.3pt] (0,0.000) -- (5.6,0.000);
\node[anchor=east,text=vzInkTwo,inner sep=1.5pt] at (0,0.000) {25};
\draw[vzGrid,line width=0.3pt] (0,0.667) -- (5.6,0.667);
\node[anchor=east,text=vzInkTwo,inner sep=1.5pt] at (0,0.667) {30};
\draw[vzGrid,line width=0.3pt] (0,1.333) -- (5.6,1.333);
\node[anchor=east,text=vzInkTwo,inner sep=1.5pt] at (0,1.333) {35};
\draw[vzGrid,line width=0.3pt] (0,2.000) -- (5.6,2.000);
\node[anchor=east,text=vzInkTwo,inner sep=1.5pt] at (0,2.000) {40};
\draw[vzGrid,line width=0.3pt] (0,2.667) -- (5.6,2.667);
\node[anchor=east,text=vzInkTwo,inner sep=1.5pt] at (0,2.667) {45};
\draw[vzGrid,line width=0.3pt] (0,3.333) -- (5.6,3.333);
\node[anchor=east,text=vzInkTwo,inner sep=1.5pt] at (0,3.333) {50};
\draw[vzAxis,line width=0.4pt] (0,0) -- (5.6,0);
\draw[vzAxis,line width=0.4pt] (0.207,0) -- (0.207,-0.06);
\node[anchor=north,text=vzInkTwo,inner sep=1.5pt] at (0.207,-0.06) {0};
\draw[vzAxis,line width=0.4pt] (1.504,0) -- (1.504,-0.06);
\node[anchor=north,text=vzInkTwo,inner sep=1.5pt] at (1.504,-0.06) {0.25};
\draw[vzAxis,line width=0.4pt] (2.800,0) -- (2.800,-0.06);
\node[anchor=north,text=vzInkTwo,inner sep=1.5pt] at (2.800,-0.06) {0.5};
\draw[vzAxis,line width=0.4pt] (4.096,0) -- (4.096,-0.06);
\node[anchor=north,text=vzInkTwo,inner sep=1.5pt] at (4.096,-0.06) {0.75};
\draw[vzAxis,line width=0.4pt] (5.393,0) -- (5.393,-0.06);
\node[anchor=north,text=vzInkTwo,inner sep=1.5pt] at (5.393,-0.06) {1};
\node[anchor=north,text=vzInk] at (2.800,-0.36) {Correction strength $\lambda$};
\node[anchor=south,rotate=90,text=vzInk] at (-0.42,1.800) {Held-out TinyGSM accuracy (\%)};
\node[anchor=south,text=vzInk,font=\small] at (2.800,3.940) {DiT};
\draw[vzMuted,line width=0.6pt,dash pattern=on 1.5pt off 1.5pt] (1.504,0) -- (1.504,3.600);
\node[anchor=south,text=vzInkTwo,inner sep=1pt] at (1.504,3.600) {$\lambda^*=0.25$};
\draw[vzT4,line width=1.1pt,line join=round,line cap=round] (0.207,2.366) -- (1.504,2.513) -- (2.800,2.576) -- (4.096,2.597) -- (5.393,2.534);
\draw[vzT2,line width=1.1pt,line join=round,line cap=round] (0.207,2.513) -- (1.504,2.871) -- (2.800,2.955) -- (4.096,2.871) -- (5.393,2.534);
\draw[vzT1,line width=1.1pt,line join=round,line cap=round] (0.207,2.702) -- (1.504,3.312) -- (2.800,2.871) -- (4.096,2.072) -- (5.393,0.473);
\draw[vzInk,line width=1.6pt,line join=round,line cap=round] (0.207,2.527) -- (1.504,2.899) -- (2.800,2.801) -- (4.096,2.513) -- (5.393,1.847);
\filldraw[fill=vzT4,draw=white,line width=0.5pt] (0.207,2.366) circle (0.065);
\filldraw[fill=vzT4,draw=white,line width=0.5pt] (1.504,2.513) circle (0.065);
\filldraw[fill=vzT4,draw=white,line width=0.5pt] (2.800,2.576) circle (0.065);
\filldraw[fill=vzT4,draw=white,line width=0.5pt] (4.096,2.597) circle (0.065);
\filldraw[fill=vzT4,draw=white,line width=0.5pt] (5.393,2.534) circle (0.065);
\filldraw[fill=vzT2,draw=white,line width=0.5pt] (0.150,2.456) rectangle (0.265,2.570);
\filldraw[fill=vzT2,draw=white,line width=0.5pt] (1.447,2.813) rectangle (1.561,2.928);
\filldraw[fill=vzT2,draw=white,line width=0.5pt] (2.743,2.898) rectangle (2.857,3.012);
\filldraw[fill=vzT2,draw=white,line width=0.5pt] (4.039,2.813) rectangle (4.153,2.928);
\filldraw[fill=vzT2,draw=white,line width=0.5pt] (5.335,2.477) rectangle (5.450,2.591);
\filldraw[fill=vzT1,draw=white,line width=0.5pt] (0.207,2.784) -- (0.137,2.662) -- (0.278,2.662) -- cycle;
\filldraw[fill=vzT1,draw=white,line width=0.5pt] (1.504,3.394) -- (1.433,3.272) -- (1.574,3.272) -- cycle;
\filldraw[fill=vzT1,draw=white,line width=0.5pt] (2.800,2.952) -- (2.730,2.830) -- (2.870,2.830) -- cycle;
\filldraw[fill=vzT1,draw=white,line width=0.5pt] (4.096,2.153) -- (4.026,2.031) -- (4.167,2.031) -- cycle;
\filldraw[fill=vzT1,draw=white,line width=0.5pt] (5.393,0.554) -- (5.322,0.433) -- (5.463,0.433) -- cycle;
\filldraw[fill=vzInk,draw=white,line width=0.5pt] (0.207,2.608) -- (0.289,2.527) -- (0.207,2.446) -- (0.126,2.527) -- cycle;
\filldraw[fill=vzInk,draw=white,line width=0.5pt] (1.504,2.980) -- (1.585,2.899) -- (1.504,2.817) -- (1.422,2.899) -- cycle;
\filldraw[fill=vzInk,draw=white,line width=0.5pt] (2.800,2.882) -- (2.881,2.801) -- (2.800,2.719) -- (2.719,2.801) -- cycle;
\filldraw[fill=vzInk,draw=white,line width=0.5pt] (4.096,2.594) -- (4.178,2.513) -- (4.096,2.432) -- (4.015,2.513) -- cycle;
\filldraw[fill=vzInk,draw=white,line width=0.5pt] (5.393,1.928) -- (5.474,1.847) -- (5.393,1.766) -- (5.311,1.847) -- cycle;
\end{tikzpicture}
\end{minipage}
\caption{Held-out selection of $\lambda$ for BCT-L (634 held-out TinyGSM problems, \sflmtft{}, seed 1). Thin lines: accuracy at each selection temperature; thick line: their mean, the selection score. The dotted line marks the selected $\lambda$. $\lambda=0$ applies no Bayes correction and $\lambda=1$ the full correction. Values in Table~\ref{tab:app-lambda}.}
\label{fig:app-lambda}
\end{figure}

\textbf{Protocol.}
For each backbone, the sweep uses the \sflmtft{} checkpoint, generation seed 1 and the GSM8K sampling protocol. Each $\lambda\in\{0,0.25,0.5,0.75,1\}$ is scored by its held-out accuracy averaged over $T\in\{0.4,0.2,0.1\}$; the maximizer, with ties broken toward the larger $\lambda$, is used at every temperature. This selects $\lambda=0.5$ for S-arch and $\lambda=0.25$ for DiT (Figure~\ref{fig:app-lambda}, Table~\ref{tab:app-lambda}).

\begin{table}[ht]
\caption{Held-out $\lambda$ sweep: correct answers out of 634 at each selection temperature and the selection score (mean accuracy over the three temperatures, \%); bold: selected value. The last column is a DiT TFT checkpoint after 500 updates, not used for selection.}
\label{tab:app-lambda}
\centering
\small
\begin{tabular}{@{}lcccccccccc@{}}
\toprule
& \multicolumn{4}{c}{S-arch} & \multicolumn{4}{c}{DiT} & DiT (500 upd.) \\
\cmidrule(lr){2-5}\cmidrule(lr){6-9}\cmidrule(l){10-10}
$\lambda$ & $T{=}0.4$ & $0.2$ & $0.1$ & Score & $T{=}0.4$ & $0.2$ & $0.1$ & Score & Score \\
\midrule
0 & 279 & 285 & 282 & 44.48 & 271 & 278 & 287 & 43.95 & 42.27 \\
0.25 & 279 & 292 & 290 & 45.27 & 278 & 295 & 316 & \textbf{46.74} & \textbf{48.74} \\
0.5 & 283 & 304 & 303 & \textbf{46.79} & 281 & 299 & 295 & 46.00 & 48.32 \\
0.75 & 284 & 305 & 295 & 46.48 & 282 & 295 & 257 & 43.85 & 45.79 \\
1 & 286 & 300 & 289 & 46.00 & 279 & 279 & 181 & 38.85 & 43.22 \\
 
\bottomrule
\end{tabular}
\end{table}

\clearpage
\section{Pass@$K$ on S-arch}
\label{app:passk}

This appendix gives the setup and values behind Section~\ref{sec:exp-passk}.

\textbf{Setup.}
We compare \sflmcont{} with standard tempering and \sflmtft{} with BCT-L ($\lambda=0.5$) on S-arch, using the checkpoints, sampler, prompts and grader of Table~\ref{tab:main} at the same eight temperatures. The sampler draws one generation per problem, so the eight generations per problem come from eight runs with generation seeds $1+10j$, $j=0,\dots,7$; the first three are the seeds of Table~\ref{tab:main} and reproduce its generations exactly. For a problem with $c$ correct generations out of $n=8$, we use the unbiased estimator $1-\binom{n-c}{k}/\binom{n}{k}$ \citep{chen2021codex}, averaged over the 1319 problems. Intervals are 95\% percentile bootstrap intervals over problems (10{,}000 resamples), paired for differences.

\begin{table}[ht]
\caption{Pass@$K$ (\%) of Figure~\ref{fig:passk}: unbiased estimator over $n=8$ generations per problem (generation seeds $1,11,\dots,71$), S-arch, 1319 GSM8K problems. Pass@1 is the mean single-generation accuracy over the eight draws and agrees with the three-seed mean of Table~\ref{tab:main} to within 0.6 points.}
\label{tab:app-passk}
\centering
\small
\begin{tabular}{@{}lcccccccc@{}}
\toprule
& \multicolumn{4}{c}{\sflmcont{}} & \multicolumn{4}{c}{\sflmtft{}} \\
& \multicolumn{4}{c}{standard tempering} & \multicolumn{4}{c}{BCT-L} \\
\cmidrule(lr){2-5}\cmidrule(l){6-9}
$T$ & $K=1$ & $2$ & $4$ & $8$ & $K=1$ & $2$ & $4$ & $8$ \\
\midrule
1 & 12.3 & 19.6 & 28.3 & 37.5 & 12.7 & 20.2 & 29.0 & 37.7 \\
0.8 & 14.4 & 22.5 & 32.1 & 42.0 & 16.5 & 25.1 & 34.7 & 44.5 \\
0.6 & 15.8 & 24.3 & 34.1 & 44.5 & 20.2 & 29.8 & 40.3 & 51.3 \\
0.5 & 16.9 & 25.8 & 36.0 & 46.8 & 21.8 & 31.3 & 41.2 & 51.1 \\
0.4 & 17.4 & 26.3 & 36.7 & 47.5 & 23.3 & 32.9 & 42.8 & 52.5 \\
0.3 & 17.6 & 26.7 & 36.9 & 47.3 & 24.4 & 33.7 & 43.2 & 52.5 \\
0.2 & 17.7 & 26.6 & 36.6 & 47.2 & 25.3 & 34.5 & 43.4 & 51.9 \\
0.1 & 17.8 & 26.8 & 37.0 & 47.5 & 27.1 & 35.7 & 43.9 & 51.1 \\
 
\bottomrule
\end{tabular}
\end{table}

\textbf{Differences.}
For every $T<1$ and $K\le8$, the paired interval of the difference (BCT-L minus standard tempering) lies above zero. At $T=0.1$ the difference is 9.2 [8.1, 10.3] points at $K=1$ and 3.6 [1.4, 5.8] at $K=8$. The number of problems solved by at least one of the eight generations is 693 and 674 for BCT-L at $T=0.3$ and $0.1$, against 624 and 627 for standard tempering.

\clearpage
\section{Standard temperature scaling on discrete diffusion baselines}
\label{app:dlm-temp}

We sweep the sampling temperature of the MDLM \citep{sahoo2024simple} and Duo \citep{sahoo2025the} TinyGSM checkpoints released by \citet{hypersphericalflows2026}, without any training or tuning. Both use the ancestral samplers of that release: at every step the denoiser logits are divided by $T$ before the softmax, and the tempered clean-token distribution enters the unchanged reverse transition (absorbing state for MDLM, uniform state for Duo). At $T=1$ this reproduces the released sampler exactly (1319 of 1319 identical programs for both models, checked with 32 steps). We use 1024 steps, generation seed 1 and the GSM8K evaluation of Section~\ref{sec:exp-setup}; the \sflm{} values are the standard-tempering results at seed 1.

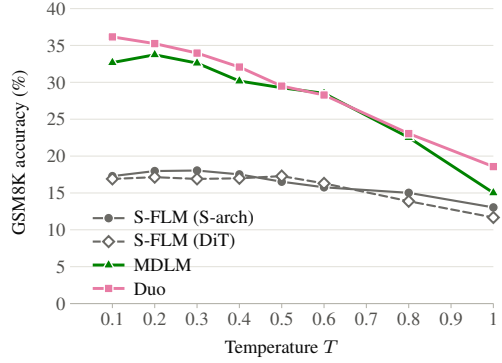
\begin{figure}[ht]
\centering
\begin{tikzpicture}[x=1cm,y=1cm,font=\scriptsize]
\definecolor{vzBlue}{HTML}{2A78D6}
\definecolor{vzOrange}{HTML}{EB6834}
\definecolor{vzAqua}{HTML}{1BAF7A}
\definecolor{vzBase}{HTML}{6B6A66}
\definecolor{vzMuted}{HTML}{9A9892}
\definecolor{vzGrid}{HTML}{E1E0D9}
\definecolor{vzAxis}{HTML}{B5B4AC}
\definecolor{vzInk}{HTML}{0B0B0B}
\definecolor{vzInkTwo}{HTML}{52514E}
\definecolor{vzViolet}{HTML}{4A3AA7}
\definecolor{vzT4}{HTML}{86B6EF}
\definecolor{vzT2}{HTML}{2A78D6}
\definecolor{vzT1}{HTML}{104281}
\definecolor{vzGreen}{HTML}{008300}
\definecolor{vzMagenta}{HTML}{E87BA4}
\draw[vzGrid,line width=0.3pt] (0,0.000) -- (5.6,0.000);
\node[anchor=east,text=vzInkTwo,inner sep=1.5pt] at (0,0.000) {0};
\draw[vzGrid,line width=0.3pt] (0,0.488) -- (5.6,0.488);
\node[anchor=east,text=vzInkTwo,inner sep=1.5pt] at (0,0.488) {5};
\draw[vzGrid,line width=0.3pt] (0,0.976) -- (5.6,0.976);
\node[anchor=east,text=vzInkTwo,inner sep=1.5pt] at (0,0.976) {10};
\draw[vzGrid,line width=0.3pt] (0,1.463) -- (5.6,1.463);
\node[anchor=east,text=vzInkTwo,inner sep=1.5pt] at (0,1.463) {15};
\draw[vzGrid,line width=0.3pt] (0,1.951) -- (5.6,1.951);
\node[anchor=east,text=vzInkTwo,inner sep=1.5pt] at (0,1.951) {20};
\draw[vzGrid,line width=0.3pt] (0,2.439) -- (5.6,2.439);
\node[anchor=east,text=vzInkTwo,inner sep=1.5pt] at (0,2.439) {25};
\draw[vzGrid,line width=0.3pt] (0,2.927) -- (5.6,2.927);
\node[anchor=east,text=vzInkTwo,inner sep=1.5pt] at (0,2.927) {30};
\draw[vzGrid,line width=0.3pt] (0,3.415) -- (5.6,3.415);
\node[anchor=east,text=vzInkTwo,inner sep=1.5pt] at (0,3.415) {35};
\draw[vzGrid,line width=0.3pt] (0,3.902) -- (5.6,3.902);
\node[anchor=east,text=vzInkTwo,inner sep=1.5pt] at (0,3.902) {40};
\draw[vzAxis,line width=0.4pt] (0,0) -- (5.6,0);
\draw[vzAxis,line width=0.4pt] (0.560,0) -- (0.560,-0.06);
\node[anchor=north,text=vzInkTwo,inner sep=1.5pt] at (0.560,-0.06) {0.1};
\draw[vzAxis,line width=0.4pt] (1.120,0) -- (1.120,-0.06);
\node[anchor=north,text=vzInkTwo,inner sep=1.5pt] at (1.120,-0.06) {0.2};
\draw[vzAxis,line width=0.4pt] (1.680,0) -- (1.680,-0.06);
\node[anchor=north,text=vzInkTwo,inner sep=1.5pt] at (1.680,-0.06) {0.3};
\draw[vzAxis,line width=0.4pt] (2.240,0) -- (2.240,-0.06);
\node[anchor=north,text=vzInkTwo,inner sep=1.5pt] at (2.240,-0.06) {0.4};
\draw[vzAxis,line width=0.4pt] (2.800,0) -- (2.800,-0.06);
\node[anchor=north,text=vzInkTwo,inner sep=1.5pt] at (2.800,-0.06) {0.5};
\draw[vzAxis,line width=0.4pt] (3.360,0) -- (3.360,-0.06);
\node[anchor=north,text=vzInkTwo,inner sep=1.5pt] at (3.360,-0.06) {0.6};
\draw[vzAxis,line width=0.4pt] (3.920,0) -- (3.920,-0.06);
\node[anchor=north,text=vzInkTwo,inner sep=1.5pt] at (3.920,-0.06) {0.7};
\draw[vzAxis,line width=0.4pt] (4.480,0) -- (4.480,-0.06);
\node[anchor=north,text=vzInkTwo,inner sep=1.5pt] at (4.480,-0.06) {0.8};
\draw[vzAxis,line width=0.4pt] (5.040,0) -- (5.040,-0.06);
\node[anchor=north,text=vzInkTwo,inner sep=1.5pt] at (5.040,-0.06) {0.9};
\draw[vzAxis,line width=0.4pt] (5.600,0) -- (5.600,-0.06);
\node[anchor=north,text=vzInkTwo,inner sep=1.5pt] at (5.600,-0.06) {1};
\node[anchor=north,text=vzInk] at (2.800,-0.36) {Temperature $T$};
\node[anchor=south,rotate=90,text=vzInk] at (-0.42,2.000) {GSM8K accuracy (\%)};
\draw[vzBase,line width=0.8pt,line join=round,line cap=round] (0.560,1.686) -- (1.120,1.753) -- (1.680,1.760) -- (2.240,1.709) -- (2.800,1.612) -- (3.360,1.538) -- (4.480,1.465) -- (5.600,1.272);
\draw[vzBase,line width=0.8pt,line join=round,line cap=round,dash pattern=on 3pt off 1.6pt] (0.560,1.649) -- (1.120,1.672) -- (1.680,1.649) -- (2.240,1.657) -- (2.800,1.686) -- (3.360,1.590) -- (4.480,1.354) -- (5.600,1.139);
\draw[vzGreen,line width=1.1pt,line join=round,line cap=round] (0.560,3.188) -- (1.120,3.291) -- (1.680,3.181) -- (2.240,2.944) -- (2.800,2.855) -- (3.360,2.781) -- (4.480,2.197) -- (5.600,1.465);
\draw[vzMagenta,line width=1.1pt,line join=round,line cap=round] (0.560,3.528) -- (1.120,3.439) -- (1.680,3.314) -- (2.240,3.129) -- (2.800,2.877) -- (3.360,2.759) -- (4.480,2.249) -- (5.600,1.812);
\filldraw[fill=vzBase,draw=white,line width=0.5pt] (0.560,1.686) circle (0.065);
\filldraw[fill=vzBase,draw=white,line width=0.5pt] (1.120,1.753) circle (0.065);
\filldraw[fill=vzBase,draw=white,line width=0.5pt] (1.680,1.760) circle (0.065);
\filldraw[fill=vzBase,draw=white,line width=0.5pt] (2.240,1.709) circle (0.065);
\filldraw[fill=vzBase,draw=white,line width=0.5pt] (2.800,1.612) circle (0.065);
\filldraw[fill=vzBase,draw=white,line width=0.5pt] (3.360,1.538) circle (0.065);
\filldraw[fill=vzBase,draw=white,line width=0.5pt] (4.480,1.465) circle (0.065);
\filldraw[fill=vzBase,draw=white,line width=0.5pt] (5.600,1.272) circle (0.065);
\filldraw[fill=white,draw=vzBase,line width=0.75pt] (0.560,1.723) -- (0.633,1.649) -- (0.560,1.576) -- (0.487,1.649) -- cycle;
\filldraw[fill=white,draw=vzBase,line width=0.75pt] (1.120,1.745) -- (1.193,1.672) -- (1.120,1.599) -- (1.047,1.672) -- cycle;
\filldraw[fill=white,draw=vzBase,line width=0.75pt] (1.680,1.723) -- (1.753,1.649) -- (1.680,1.576) -- (1.607,1.649) -- cycle;
\filldraw[fill=white,draw=vzBase,line width=0.75pt] (2.240,1.730) -- (2.313,1.657) -- (2.240,1.584) -- (2.167,1.657) -- cycle;
\filldraw[fill=white,draw=vzBase,line width=0.75pt] (2.800,1.760) -- (2.873,1.686) -- (2.800,1.613) -- (2.727,1.686) -- cycle;
\filldraw[fill=white,draw=vzBase,line width=0.75pt] (3.360,1.663) -- (3.433,1.590) -- (3.360,1.517) -- (3.287,1.590) -- cycle;
\filldraw[fill=white,draw=vzBase,line width=0.75pt] (4.480,1.427) -- (4.553,1.354) -- (4.480,1.280) -- (4.407,1.354) -- cycle;
\filldraw[fill=white,draw=vzBase,line width=0.75pt] (5.600,1.212) -- (5.673,1.139) -- (5.600,1.066) -- (5.527,1.139) -- cycle;
\filldraw[fill=vzGreen,draw=white,line width=0.5pt] (0.560,3.269) -- (0.490,3.147) -- (0.630,3.147) -- cycle;
\filldraw[fill=vzGreen,draw=white,line width=0.5pt] (1.120,3.373) -- (1.050,3.251) -- (1.190,3.251) -- cycle;
\filldraw[fill=vzGreen,draw=white,line width=0.5pt] (1.680,3.262) -- (1.610,3.140) -- (1.750,3.140) -- cycle;
\filldraw[fill=vzGreen,draw=white,line width=0.5pt] (2.240,3.025) -- (2.170,2.903) -- (2.310,2.903) -- cycle;
\filldraw[fill=vzGreen,draw=white,line width=0.5pt] (2.800,2.936) -- (2.730,2.814) -- (2.870,2.814) -- cycle;
\filldraw[fill=vzGreen,draw=white,line width=0.5pt] (3.360,2.862) -- (3.290,2.740) -- (3.430,2.740) -- cycle;
\filldraw[fill=vzGreen,draw=white,line width=0.5pt] (4.480,2.278) -- (4.410,2.156) -- (4.550,2.156) -- cycle;
\filldraw[fill=vzGreen,draw=white,line width=0.5pt] (5.600,1.546) -- (5.530,1.424) -- (5.670,1.424) -- cycle;
\filldraw[fill=vzMagenta,draw=white,line width=0.5pt] (0.503,3.471) rectangle (0.617,3.585);
\filldraw[fill=vzMagenta,draw=white,line width=0.5pt] (1.063,3.382) rectangle (1.177,3.497);
\filldraw[fill=vzMagenta,draw=white,line width=0.5pt] (1.623,3.256) rectangle (1.737,3.371);
\filldraw[fill=vzMagenta,draw=white,line width=0.5pt] (2.183,3.072) rectangle (2.297,3.186);
\filldraw[fill=vzMagenta,draw=white,line width=0.5pt] (2.743,2.820) rectangle (2.857,2.934);
\filldraw[fill=vzMagenta,draw=white,line width=0.5pt] (3.303,2.702) rectangle (3.417,2.816);
\filldraw[fill=vzMagenta,draw=white,line width=0.5pt] (4.423,2.191) rectangle (4.537,2.306);
\filldraw[fill=vzMagenta,draw=white,line width=0.5pt] (5.543,1.755) rectangle (5.657,1.869);
\draw[vzMagenta,line width=1.1pt] (0.300,0.200) -- (0.720,0.200);
\filldraw[fill=vzMagenta,draw=white,line width=0.5pt] (0.453,0.143) rectangle (0.567,0.257);
\node[anchor=west,text=vzInk,inner sep=1pt] at (0.800,0.200) {Duo};
\draw[vzGreen,line width=1.1pt] (0.300,0.510) -- (0.720,0.510);
\filldraw[fill=vzGreen,draw=white,line width=0.5pt] (0.510,0.591) -- (0.440,0.469) -- (0.580,0.469) -- cycle;
\node[anchor=west,text=vzInk,inner sep=1pt] at (0.800,0.510) {MDLM};
\draw[vzBase,line width=0.8pt,dash pattern=on 3pt off 1.6pt] (0.300,0.820) -- (0.720,0.820);
\filldraw[fill=white,draw=vzBase,line width=0.75pt] (0.510,0.893) -- (0.583,0.820) -- (0.510,0.747) -- (0.437,0.820) -- cycle;
\node[anchor=west,text=vzInk,inner sep=1pt] at (0.800,0.820) {S-FLM (DiT)};
\draw[vzBase,line width=0.8pt] (0.300,1.130) -- (0.720,1.130);
\filldraw[fill=vzBase,draw=white,line width=0.5pt] (0.510,1.130) circle (0.065);
\node[anchor=west,text=vzInk,inner sep=1pt] at (0.800,1.130) {S-FLM (S-arch)};
\end{tikzpicture}
\caption{Standard temperature scaling on the TinyGSM baselines released with \sflm{} (NFE $=1024$, seed 1). Gray: \sflm{} (Section~\ref{sec:exp-main}). Values in Table~\ref{tab:app-dlm-temp}.}
\label{fig:app-dlm-temp}
\end{figure}

\begin{table}[ht]
\caption{Accuracy (\%) and correct answers out of 1319 for Figure~\ref{fig:app-dlm-temp}. \sflm{} uses the exact velocity.}
\label{tab:app-dlm-temp}
\centering
\small
\begin{tabular}{@{}lcccc@{}}
\toprule
$T$ & MDLM & Duo & \sflm{} (S-arch) & \sflm{} (DiT) \\
\midrule
1 & 15.0 (198) & 18.6 (245) & 13.0 (172) & 11.7 (154) \\
0.8 & 22.5 (297) & 23.0 (304) & 15.0 (198) & 13.9 (183) \\
0.6 & 28.5 (376) & 28.3 (373) & 15.8 (208) & 16.3 (215) \\
0.5 & 29.3 (386) & 29.5 (389) & 16.5 (218) & 17.3 (228) \\
0.4 & 30.2 (398) & 32.1 (423) & 17.5 (231) & 17.0 (224) \\
0.3 & 32.6 (430) & 34.0 (448) & 18.0 (238) & 16.9 (223) \\
0.2 & 33.7 (445) & 35.3 (465) & 18.0 (237) & 17.1 (226) \\
0.1 & 32.7 (431) & 36.2 (477) & 17.3 (228) & 16.9 (223) \\
 
\bottomrule
\end{tabular}
\end{table}

At $T=0.1$, MDLM and Duo solve 32.7\% and 36.2\%, in line with the 33\% and 36\% reported by \citet{hypersphericalflows2026}; at $T=1$ they solve 15.0\% and 18.6\%, against the reported 18.0\% and 17.2\%.

\clearpage
\section{Sudoku-Hard}
\label{app:sudoku}

This appendix gives the setup and values behind Section~\ref{sec:exp-sudoku}.

\textbf{Task.}
We use the Sudoku-Hard task and puzzle generator of \sflm{} \citep{hypersphericalflows2026} unchanged. Every puzzle has 30 clues and a unique solution (checked with a solution counter), and the 48{,}000 training and 2{,}000 test puzzles share no puzzle or solution. A sequence is the puzzle followed by its solution, each written as 81 digits with row separators (a vocabulary of 12 tokens). The puzzle is a fixed prefix during sampling, and a test puzzle counts as solved only if the generated solution equals the reference.

\textbf{Model and training.}
We train with the released Sudoku recipe of \sflm{} unchanged: a DiT denoiser with 8 blocks and hidden size 512, the geodesic bridge with its truncated adaptive noise schedule, 20{,}000 updates with a batch of 256 puzzles on one GPU, AdamW with a constant learning rate of $3\times10^{-4}$ after warmup, EMA decay 0.9999, and training seed 1. Our code reproduces the released code bit for bit on this recipe (weights, EMA weights and optimizer state after 1{,}200 updates). We call the result the base checkpoint. \sflmcont{} and \sflmtft{} resume its complete training state for 2{,}000 updates; \sflmtft{} uses the TFT setting of Section~\ref{sec:exp-setup} with $T$-conditioning. Each checkpoint is a single training run.

\textbf{Evaluation.}
All samplers use the released Sudoku sampler (180 NFE, exact velocity, argmax decoding, EMA weights, batches of 64 puzzles on one GPU) with generation seeds 1, 11 and 21; only the temperature and the BCT rule change between runs, and for a given seed all samplers start from the same noise. At $T=1$ the training-free samplers coincide up to floating-point differences, which change the number of solved puzzles by at most 2 per seed.

\textbf{Selection of $\lambda$.}
We select $\lambda$ with the protocol of Appendix~\ref{app:practical} on 2{,}000 new puzzles from the same generator with a different seed, disjoint from the training and test puzzles. Held-out accuracy increases with $\lambda$ at every selection temperature (Table~\ref{tab:app-sudoku-lambda}), so $\lambda=1$.

\begin{table}[ht]
\caption{Sudoku-Hard exact solve rate (\%), mean $\pm$ standard deviation over generation seeds 1, 11 and 21 (2{,}000 test puzzles, NFE $=180$). BCT-L uses the held-out $\lambda=1$. The seed standard deviation is dominated by a shift shared by all samplers.}
\label{tab:app-sudoku}
\centering
\small
\setlength{\tabcolsep}{4pt}
\begin{tabular}{@{}lccccccc@{}}
\toprule
& \multicolumn{3}{c}{Base checkpoint} & \multicolumn{3}{c}{\sflmcont{}} & \sflmtft{} \\
\cmidrule(lr){2-4}\cmidrule(lr){5-7}\cmidrule(l){8-8}
$T$ & Standard & BCT-E & BCT-P & Standard & BCT-E & BCT-P & BCT-L \\
\midrule
1 & 42.9 $\pm$ 1.5 & 42.9 $\pm$ 1.6 & 42.9 $\pm$ 1.5 & 41.7 $\pm$ 1.2 & 41.7 $\pm$ 1.2 & 41.7 $\pm$ 1.2 & 41.7 $\pm$ 1.1 \\
0.8 & 42.6 $\pm$ 1.6 & 43.7 $\pm$ 1.4 & 43.6 $\pm$ 1.0 & 41.6 $\pm$ 1.2 & 42.6 $\pm$ 1.2 & 42.4 $\pm$ 1.0 & 42.7 $\pm$ 0.7 \\
0.6 & 42.3 $\pm$ 1.3 & 44.1 $\pm$ 1.2 & 44.2 $\pm$ 1.1 & 41.4 $\pm$ 1.1 & 43.1 $\pm$ 1.0 & 42.7 $\pm$ 1.2 & 44.3 $\pm$ 1.2 \\
0.5 & 42.2 $\pm$ 1.3 & 44.4 $\pm$ 1.3 & 44.4 $\pm$ 1.2 & 41.3 $\pm$ 1.0 & 43.3 $\pm$ 1.2 & 43.0 $\pm$ 0.8 & 45.1 $\pm$ 1.3 \\
0.4 & 42.1 $\pm$ 1.2 & 44.8 $\pm$ 1.4 & 44.8 $\pm$ 1.1 & 41.2 $\pm$ 1.0 & 43.5 $\pm$ 1.2 & 43.2 $\pm$ 0.5 & 45.6 $\pm$ 1.3 \\
0.3 & 42.0 $\pm$ 1.2 & 45.0 $\pm$ 1.4 & 44.7 $\pm$ 0.9 & 41.2 $\pm$ 1.0 & 43.7 $\pm$ 1.2 & 43.3 $\pm$ 0.5 & 45.9 $\pm$ 1.1 \\
0.2 & 41.9 $\pm$ 1.3 & 45.2 $\pm$ 1.4 & 44.8 $\pm$ 0.4 & 41.2 $\pm$ 1.0 & 44.0 $\pm$ 1.2 & 43.6 $\pm$ 0.1 & 46.4 $\pm$ 1.1 \\
0.1 & 41.7 $\pm$ 1.3 & 45.5 $\pm$ 1.4 & 44.9 $\pm$ 0.4 & 41.0 $\pm$ 1.0 & 44.2 $\pm$ 1.4 & 43.7 $\pm$ 0.5 & 46.8 $\pm$ 1.2 \\
 
\bottomrule
\end{tabular}
\end{table}

\begin{table}[ht]
\caption{Held-out selection of $\lambda$ for BCT-L on Sudoku-Hard: exact solve rate (\%) of \sflmtft{} on 2{,}000 held-out puzzles (generation seed 1, NFE $=180$). Mean: the selection score. Bold: selected. No test puzzle was used.}
\label{tab:app-sudoku-lambda}
\centering
\small
\begin{tabular}{@{}lcccc@{}}
\toprule
$\lambda$ & $T=0.4$ & $T=0.2$ & $T=0.1$ & Mean \\
\midrule
0 & 41.8 & 40.8 & 40.1 & 40.9 \\
0.25 & 42.8 & 43.5 & 43.3 & 43.2 \\
0.5 & 44.0 & 44.5 & 44.9 & 44.5 \\
0.75 & 44.9 & 45.4 & 46.9 & 45.7 \\
\textbf{1} & \textbf{45.6} & \textbf{46.3} & \textbf{47.5} & \textbf{46.5} \\
 
\bottomrule
\end{tabular}
\end{table}

\section{BCT for vMF flows}
\label{app:sfcd}

This appendix gives the instantiation and setup of Section~\ref{sec:exp-sfcd}. SFCD \citep{sphericalflows2026} keeps learned token embeddings on the unit sphere but uses the von Mises--Fisher bridge $q_t(\vz\mid x)=\mathrm{vMF}(\ve_x,\kappa_t)$ with increasing concentration and $\kappa_0=0$. Its sampler integrates the posterior-induced velocity \eqp{eq:marg-vel} on a time warp chosen by the release. We use the released TinyGSM checkpoint (250k updates) with its architecture, warp and sampler unchanged, and change only the velocity.

\textbf{Tempering the vMF bridge.}
The tempered bridge is $\mathrm{vMF}(\ve_x,\kappa_t/T)$ \eqp{eq:vmf-tempered}, whose normalizer does not depend on $x$, so BCT-P keeps the posterior weights of standard tempering. Because the vMF bridge is indexed by its concentration, the released conditional velocity has the form $u_t(\vz\mid x)=\dot\kappa_t\,V_{\kappa_t}(\vz\mid x)$, where $V_\kappa(\cdot\mid x)$ transports $\mathrm{vMF}(\ve_x,\kappa)$ as $\kappa$ increases. The tempered bridge is the same family at concentration $\kappa_t/T$, so its conditional velocity is
\begin{equation}
u_{t,T}(\vz\mid x)=\frac{\dot\kappa_t}{T}\,V_{\kappa_t/T}(\vz\mid x):
\label{eq:vmf-tempered-velocity}
\end{equation}
the released velocity evaluated at concentration $\kappa_t/T$, with the concentration advancing $1/T$ times as fast. BCT-P and BCT-L integrate this velocity on the warp grid of the release. BCT-E keeps the original bridge and dynamics and, since $\log q_t(\vz\mid x)=\kappa_t\langle\ve_x,\vz\rangle$ up to a constant in $x$, adds $(1-\beta)\,\kappa_t\langle\ve_x,\vz\rangle$ to the tempered logits. At $T=1$ every variant reduces to the released sampler.

\textbf{TFT and BCT-L.}
TFT uses the parameterization of Section~\ref{sec:bct-instantiations}: the network outputs a remainder $a_\theta$, the posterior is $\softmax(\bm a_\theta+\bm g)$ with the analytic term $g_x=(\kappa_t/T)\langle\ve_x,\vz\rangle$, and training uses the cross-entropy of \eqp{eq:tft}. The temperature enters the network through $\kappa_t/(T\kappa_{\max})$. BCT-L \eqp{eq:bctl-logits} then takes the form $\ell_x=\beta\,a_{\theta,x}+[\beta+\lambda(1-\beta)]\,g_x$.

\textbf{Continued training.}
Both continuations resume the released checkpoint for 20{,}000 updates: \sflmcont{} with the original recipe (with the warp frozen, as in TFT) and \sflmtft{} with the temperature distribution of Section~\ref{sec:exp-setup} and latents drawn from $q_{t,T}$. Both are evaluated with an EMA accumulated over the continuation (decay 0.9999); the released checkpoint uses its released EMA weights. Each continuation is a single training run.

\textbf{Evaluation.}
We use the released SFCD evaluation of the 1319 GSM8K test problems (warp-aware ODE sampler, NFE $=1024$, batches of 32, and its own answer checker) with generation seeds 42--46; every cell of Table~\ref{tab:app-sfcd} is complete over the five seeds. The separate $T=1$ runs of BCT-E on the released checkpoint reproduce standard tempering exactly.

\textbf{Selection of $\lambda$.}
We select $\lambda$ with the protocol of Appendix~\ref{app:practical} on 1024 TinyGSM validation problems that the continuation did not read and that share no problem with the GSM8K test set. The score increases with $\lambda$ (51.92, 53.19, 53.87, 54.07 and 54.39\% for $\lambda=0,0.25,0.5,0.75,1$), so $\lambda=1$.

\begin{table}[ht]
\caption{BCT on SFCD: GSM8K accuracy (\%), mean $\pm$ sample standard deviation over generation seeds 42--46 (1319 problems, NFE $=1024$). Released: the SFCD TinyGSM checkpoint with its released EMA weights. Cont.\ and TFT: 20{,}000 updates from it, evaluated with an adaptation-local EMA. BCT-L uses the held-out $\lambda=1$. At $T=1$ every correction is the identity, so the three columns of a model coincide.}
\label{tab:app-sfcd}
\centering
\scriptsize
\setlength{\tabcolsep}{3pt}
\begin{tabular}{@{}lccccccc@{}}
\toprule
& \multicolumn{3}{c}{Released} & \multicolumn{3}{c}{\sflmcont{}} & \sflmtft{} \\
\cmidrule(lr){2-4}\cmidrule(lr){5-7}\cmidrule(l){8-8}
$T$ & Standard & BCT-E & BCT-P & Standard & BCT-E & BCT-P & BCT-L \\
\midrule
1 & 23.55 $\pm$ 0.79 & 23.55 $\pm$ 0.79 & 23.55 $\pm$ 0.79 & 23.59 $\pm$ 0.71 & 23.59 $\pm$ 0.71 & 23.59 $\pm$ 0.71 & 21.55 $\pm$ 0.86 \\
0.8 & 27.79 $\pm$ 0.83 & 29.19 $\pm$ 1.43 & 29.39 $\pm$ 0.48 & 27.99 $\pm$ 0.27 & 29.80 $\pm$ 0.75 & 29.28 $\pm$ 0.74 & 28.19 $\pm$ 0.97 \\
0.6 & 30.34 $\pm$ 0.63 & 31.71 $\pm$ 0.64 & 31.63 $\pm$ 0.77 & 30.61 $\pm$ 0.77 & 31.54 $\pm$ 0.78 & 32.43 $\pm$ 0.45 & 32.05 $\pm$ 0.47 \\
0.5 & 30.80 $\pm$ 0.80 & 31.40 $\pm$ 0.62 & 32.36 $\pm$ 1.03 & 31.80 $\pm$ 0.77 & 31.21 $\pm$ 0.47 & 33.28 $\pm$ 0.39 & 33.54 $\pm$ 0.63 \\
0.4 & 31.19 $\pm$ 0.99 & 28.26 $\pm$ 0.52 & 33.39 $\pm$ 0.90 & 32.57 $\pm$ 0.83 & 26.37 $\pm$ 0.32 & 33.93 $\pm$ 0.58 & 34.33 $\pm$ 0.97 \\
0.3 & 31.75 $\pm$ 0.91 & 15.30 $\pm$ 0.67 & 33.63 $\pm$ 0.69 & 32.62 $\pm$ 0.61 & 2.93 $\pm$ 0.25 & 33.54 $\pm$ 0.75 & 34.47 $\pm$ 0.55 \\
0.2 & 31.71 $\pm$ 0.40 & 0.08 $\pm$ 0.11 & 32.05 $\pm$ 0.67 & 32.16 $\pm$ 0.40 & 0.02 $\pm$ 0.03 & 32.24 $\pm$ 0.28 & 33.24 $\pm$ 1.12 \\
0.1 & 30.99 $\pm$ 0.55 & 0.00 $\pm$ 0.00 & 26.63 $\pm$ 0.85 & 31.25 $\pm$ 0.42 & 0.00 $\pm$ 0.00 & 28.49 $\pm$ 0.54 & 31.68 $\pm$ 0.53 \\
 
\bottomrule
\end{tabular}
\end{table}

\end{document}